\documentclass{article} 
\usepackage{iclr2027_conference,times}

\usepackage{amsmath,amsfonts,bm}

\def\eqref#1{equation~\ref{#1}}

\def\1{\bm{1}}

\DeclareMathAlphabet{\mathsfit}{\encodingdefault}{\sfdefault}{m}{sl}
\SetMathAlphabet{\mathsfit}{bold}{\encodingdefault}{\sfdefault}{bx}{n}

\usepackage{hyperref}
\usepackage{url}

\title{One Adapter, Every Resolution: Gated Low-Rank Adaptation for Remote Sensing VLMs}

\usepackage{amsmath} 
\usepackage[ruled,vlined]{algorithm2e}
\usepackage{enumitem}
\usepackage{tabularx}

\usepackage{multirow}
\usepackage{xcolor,colortbl}
\usepackage{array}

\definecolor{rank1}{HTML}{F5C2C7}
\definecolor{rank2}{HTML}{B8D8E8}
\definecolor{rank3}{HTML}{F4E4BC}

\definecolor{headerbg}{RGB}{235,235,235}
\definecolor{groupbg}{RGB}{245,245,245}
\definecolor{backbonebg}{RGB}{248,248,248}
\definecolor{controlbg}{RGB}{235,244,252}
\definecolor{peftbg}{RGB}{244,238,250}
\definecolor{ourrow}{RGB}{232,246,232}

\usepackage[utf8]{inputenc}
\usepackage[T1]{fontenc}
\usepackage{booktabs}
\usepackage{amsfonts}
\usepackage{nicefrac}
\usepackage{microtype}
\usepackage{wrapfig}
\usepackage{graphicx}

\author{%
  Song Zhang\thanks{Equal contribution.} \\
  Nanjing University \\
  \texttt{song-zhang@smail.nju.edu.cn} \\
  \And
  Yanlong Chen\footnotemark[1]\thanks{Corresponding authors.} \\
  Nanyang Technological University \\
  \texttt{yanlong.chen@ntu.edu.sg} \\
  \And
  Yining Chen \\
  Nanjing University of Posts and Telecommunications \\
  \texttt{b24021105@njupt.edu.cn} \\
  \And
  Xiaowei Zhang \\
  Nanjing University \\
  \texttt{zhangxw@nju.edu.cn} \\
  \And
  Yawei Li\footnotemark[2] \\
  Nanyang Technological University \\
  \texttt{yawei.li@ntu.edu.sg} \\
}

\iclrfinalcopy

\begin{document}

\maketitle

\lhead{Preprint}

\begin{abstract}
Remote sensing imagery spans ground sampling distances (GSDs) from centimeters to tens of meters, so both the visual evidence for a geographic concept and the questions it can support change with physical scale. Existing remote sensing vision-language models (RS-VLMs) either ignore GSD or encode it as a text token, applying one set of scale agnostic parameters across this spectrum. We show that GSD is decodable from frozen visual features, but exploiting it requires scale dependent adaptation. We therefore introduce \textbf{ScaleEarth}, which conditions low-rank adaptation on the continuous scalar $s=\log_{10}(\mathrm{GSD})$. Its core module, \textbf{CS-HLoRA}, gates the rank dimensions of a single shared LoRA with sigmoid functions of $s$ whose thresholds are initialized at object, structure, and semantic scales; the gates route gradients by scale during training and select the active adaptation subspace at inference, with the backbone frozen. When metadata is unavailable, a heteroscedastic head, \textbf{SSE-U}, estimates $s$ with calibrated uncertainty and falls back to a default scale when uncertain. To align supervision with this mechanism, we build \textbf{GeoScale-VQA}, 1.5M QA pairs in which every sample carries a resolved GSD and newly generated questions are conditioned on the same $s$. With an 8B backbone, ScaleEarth achieves an average score of
\textbf{59.4} on XLRS-Bench ($+5.2$ over the strongest RS specialist).
On OmniEarth-Bench, it achieves an average score of \textbf{40.71}
($+7.45$ over the strongest open-source baseline). The largest gains
occur on scale-sensitive sub-tasks. The 4-bit variant can be deployed
on a single 40\,GB GPU and retains an average score of 57.4 on
XLRS-Bench, with substantially lower adaptation and deployment costs.
\end{abstract}

\section{Introduction}

In remote sensing, resolution determines not just how an object looks,
but whether it can be resolved at all. Ground sampling distance (GSD)
spans over two orders of magnitude, from $\sim$0.06,m/pixel in aerial
imagery to $\sim$10,m/pixel or coarser in satellite
products~\citep{volpi2016dense,costa2026remote}.
Coarser sampling removes fine spatial detail and changes which objects
and structures remain distinguishable, providing the minimum mapping unit
rationale behind our scale tier initialization~\citep{mas2010sensitivity}.
Vision-language models (VLMs), successful on natural
images~\citep{liu2023visual,bai2025qwen3}, are now being adapted to
remote sensing~\citep{kuckreja2024geochat,zhang2024earthgpt},
yet scale-agnostic adaptation applies the same policy across these
physically distinct regimes.
We show that GSD is decodable from frozen visual features, but
decodability alone does not ensure that adaptation uses it to guide the model's computation.
This motivates conditioning adaptation directly on GSD, so that physical
scale guides how the model processes the available evidence rather than
remaining implicit in its representation without explicitly conditioning the adaptation process.

\begin{figure}[t]
    \vspace{-0.6em}
    \centering
    \includegraphics[width=\linewidth]{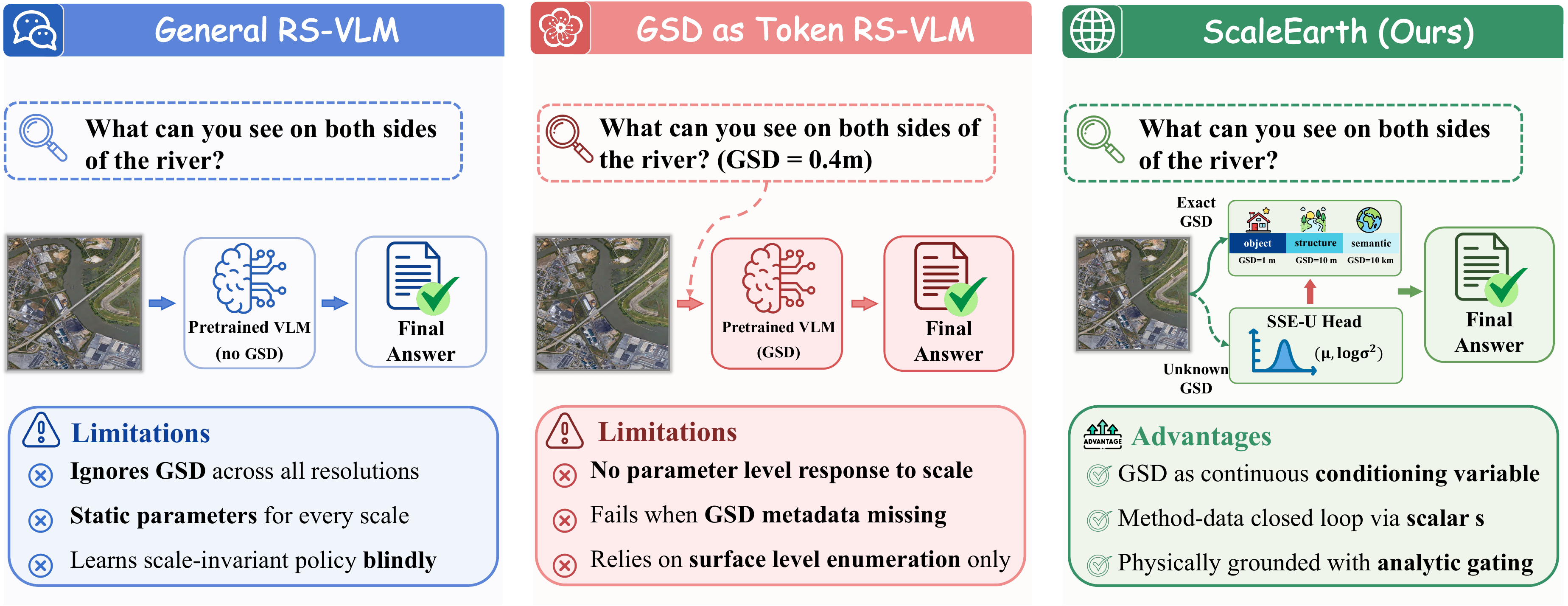}
    \vspace{-0.8em}
    \caption{\textbf{Three paradigms for handling resolution in RS-VLMs.}
    General RS-VLMs ignore GSD; GSD-as-token approaches inject it as text but elicit no parameter-level response and fail without metadata. \textbf{ScaleEarth} (ours) treats GSD as a continuous conditioning variable $s$ that drives tier-structured gating, with SSE-U predicting $(\mu, \log\sigma^2)$ when GSD is unknown.}
    \vspace{-1.0em}
    \label{fig:teaser}
\end{figure}

\vspace{-0.5em}
\paragraph{Why current RS-VLMs fall short.} As shown in the left and middle panels of Fig.~\ref{fig:teaser}, existing RS-VLMs~\citep{hu2023remote} handle scale in two limited ways. Scale agnostic finetuning ignores GSD and uses the same parameters across all resolutions, which weakens both fine-detail recognition and coarse regional reasoning. \emph{GSD-as-text-token} inserts metadata such as `0.4\,m resolution' into the prompt, but the scalar is tokenized as text, induces no parameter level adaptation, and fails when metadata is unavailable. Neither approach treats GSD as a continuous physical variable that determines which visual evidence is meaningful and which computation path should be activated. Qualitative examples in Appendix~\ref{appendix:gsd_injection_compare} show the same limitation in generated descriptions. 

This blind spot has become increasingly costly as RS benchmarks move toward real world, ultra high resolution scenarios with heterogeneous sensors. In XLRS-Bench~\citep{wang2025xlrs}, which features imagery up to $10{,}000\times10{,}000$ pixels, even state-of-the-art general VLMs such as Qwen3-VL-235B~\citep{bai2025qwen3} achieve low average accuracy. On OmniEarth-Bench, which spans multiple sensors and GSD regimes from sub-meter aerial imagery to decameter scale satellite products, no model approaches expert accuracy on any sphere (Appendix~\ref{app:omniearth_results}). Model size and data volume alone do not close this gap; in a controlled factorial (Appendix~\ref{app:factorial}), changing the adaptation mechanism contributes more than changing the corpus at a matched sample count. We therefore argue for a scale aware finetuning
recipe in which the physical resolution is no longer a passive annotation but an active signal that shapes the model's internal computation across scales.

\vspace{-0.8em}
\paragraph{Our approach: GSD as a continuous conditioning variable.}
We introduce \textbf{ScaleEarth} (Fig.~\ref{fig:teaser}, right), a
backbone preserving and parameter efficient framework built on
Qwen3-VL~\citep{bai2025qwen3} that elevates GSD from passive metadata to
a first class continuous conditioning variable. Its core component,
\textbf{CS-HLoRA}, gates the LoRA~\citep{hu2022lora} low-rank subspace
through a differentiable function of GSD, dynamically routing a shared
set of trainable adapter parameters across object, structure, and
semantic level tiers grounded in remote sensing minimum mapping unit
theory~\citep{mas2010sensitivity}. As the input GSD varies, the active LoRA dimensions shift smoothly, so that the adaptation depends on the input scale instead of a single update shared by every GSD regime, while leaving the pretrained Qwen3-VL backbone unchanged. A companion head,
\textbf{SSE-U}, predicts GSD with calibrated uncertainty from visual
features, providing a fallback conditioning signal and keeping the framework operational when sensor metadata is
missing or unreliable. On the data side, we construct
\textbf{GeoScale-VQA}, a 1.5M sample corpus in which every sample
carries the same physical scalar that drives CS-HLoRA's gating and in
which newly generated questions are conditioned on it, aligning the
model architecture and training signal around a shared notion of scale.
Sec.~\ref{sec:dataset} describes the construction of
\textsc{GeoScale-VQA}, while Sec.~\ref{sec:method} details CS-HLoRA,
SSE-U, and scale conditioned decoding. The adapters are trained on a
frozen 8B backbone; a QLoRA~\citep{dettmers2023qlora} variant with a
4-bit backbone is reported alongside as an efficiency configuration.

\vspace{-0.8em}
\paragraph{Contributions.}
\label{sec:contributions}
Our main contributions are as follows:
\vspace{-0.3em}
\begin{itemize}[leftmargin=*, nosep]
    \item We propose \textbf{CS-HLoRA}, a scale conditioned finetuning
    method that modulates the LoRA low-rank subspace through a GSD driven
    gate with physically grounded tier initialization, enabling one
    unified model to adapt its computation to physical scale without
    modifying the backbone, sensor specific retraining, or separate
    resolution specific adapters.
    \item We introduce \textbf{SSE-U}, a heteroscedastic scale estimation
    head that predicts GSD and uncertainty from visual features alone,
    reducing dependence on complete or reliable sensor metadata at
    deployment while supporting uncertainty aware scale conditioned
    decoding.
    \item We construct \textbf{GeoScale-VQA}, a scale-aware RS-VQA corpus with 1.5M samples for training resolution-conditioned remote sensing VLMs. GSD guides both the training supervision and parameter-efficient adaptation, linking the data and the method in a closed loop.
\end{itemize}

\section{Dataset}
\label{sec:dataset}

A prerequisite for scale conditioned adaptation is training data that
exposes the model to both diverse remote sensing semantics and explicit
physical scale information. We therefore construct
\textsc{GeoScale-VQA}, a scale layered RS-VQA corpus, and describe its
GSD annotation, VQA generation, and quality filtering pipeline in this
section. The full data composition, including source datasets, sample counts, GSD annotation types, and their roles in training, is reported in Appendix~\ref{app:training-recipe}, Table~\ref{tab:stage2-data}.

\subsection{GeoScale-VQA: A Scale-Layered Remote Sensing VQA Corpus}

\begin{wrapfigure}{r}{0.48\linewidth}
    \vspace{-1.2em}
    \centering
    \includegraphics[width=\linewidth]{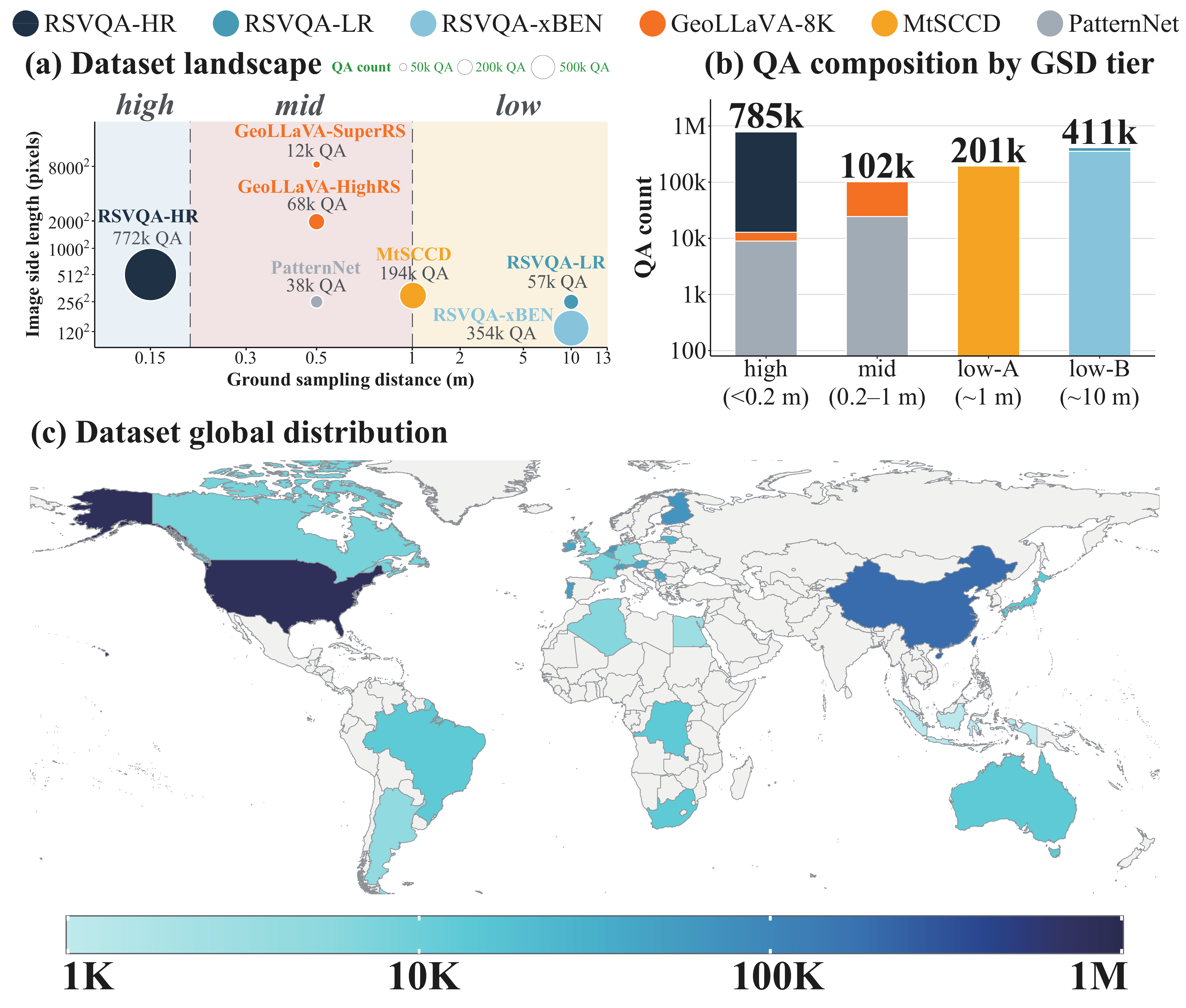}
    \caption{\textbf{GeoScale-VQA at a glance.}
    \textbf{(a)} Source-dataset landscape on the GSD $\times$
    image-size plane, where marker area encodes QA count.
    \textbf{(b)} QA composition across GSD tiers, coloured by source
    dataset.
    \textbf{(c)} Global spatial coverage of the resulting corpus.}
    \label{fig:data_landscape}
    \vspace{-1em}
\end{wrapfigure}

CS-HLoRA is motivated by a fundamental property of remote sensing imagery, namely its inherently scale dependent nature. The same object category observed at $0.1$\,m and $10$\,m ground sampling distance (GSD)
exhibits qualitatively different visual evidence, ranging from fine grained object details to coarse landscape level patterns.
This phenomenon is well established in the RS literature, where spatial resolution determines both the observable structures and the semantic
interpretation of the scene~\citep{warner2009remote,duveiller2010conceptual}.
Therefore, a training corpus for scale aware RS reasoning should not only span multiple resolutions, but also explicitly encode the physical scale
of each sample.

Existing RS-VQA corpora only partially satisfy this requirement. Some datasets are concentrated in a single GSD regime, such as RSVQA-HR at $0.15$\,m and RSVQA-LR at $10$\,m, while others omit or under-specify per-sample GSD annotations. As a result, a
VLM trained on their union cannot reliably determine at what scale
a pixel should be interpreted. To address this limitation, we build
\textbf{GeoScale-VQA}, a scale layered corpus that combines
GSD-annotated RS-VQA data with newly generated scale-aware VQA samples.
As summarized in Fig.~\ref{fig:data_landscape}, GeoScale-VQA contains
$1.5$M QA pairs spanning $0.06$--$10$\,m GSD across six
continents.

Throughout the pipeline, we organize GSD into three tiers aligned with
the CS-HLoRA anchor groups (Sec.~\ref{sec:method_stage2}), corresponding
to object, structure, and semantic level evidence; the correspondence
is ordinal rather than numerical (see Sec.~\ref{sec:method_stage2}).
Concretely, we use a high tier for $g < 0.2$\,m, a mid tier for
$0.2\,\text{m} \le g < 1.0$\,m, and a low tier for $g \ge 1.0$\,m.
Where finer reporting is useful, the low tier is split into low-A
($1.0\,\text{m} \le g < 10$\,m, metre scale sources such as MtSCCD and
coarse PatternNet classes) and low-B ($g \ge 10$\,m, decametre
Sentinel-2 based sources such as RSVQA-LR and RSVQAxBEN).
For range annotated samples
$[g_{\mathrm{lo}}, g_{\mathrm{hi}}]$, we assign the tier using the
geometric mean, the same statistic used for range samples at evaluation
time (Sec.~\ref{sec:method_stage2}).

\begin{figure}[t]
    \centering
    \includegraphics[width=\linewidth]{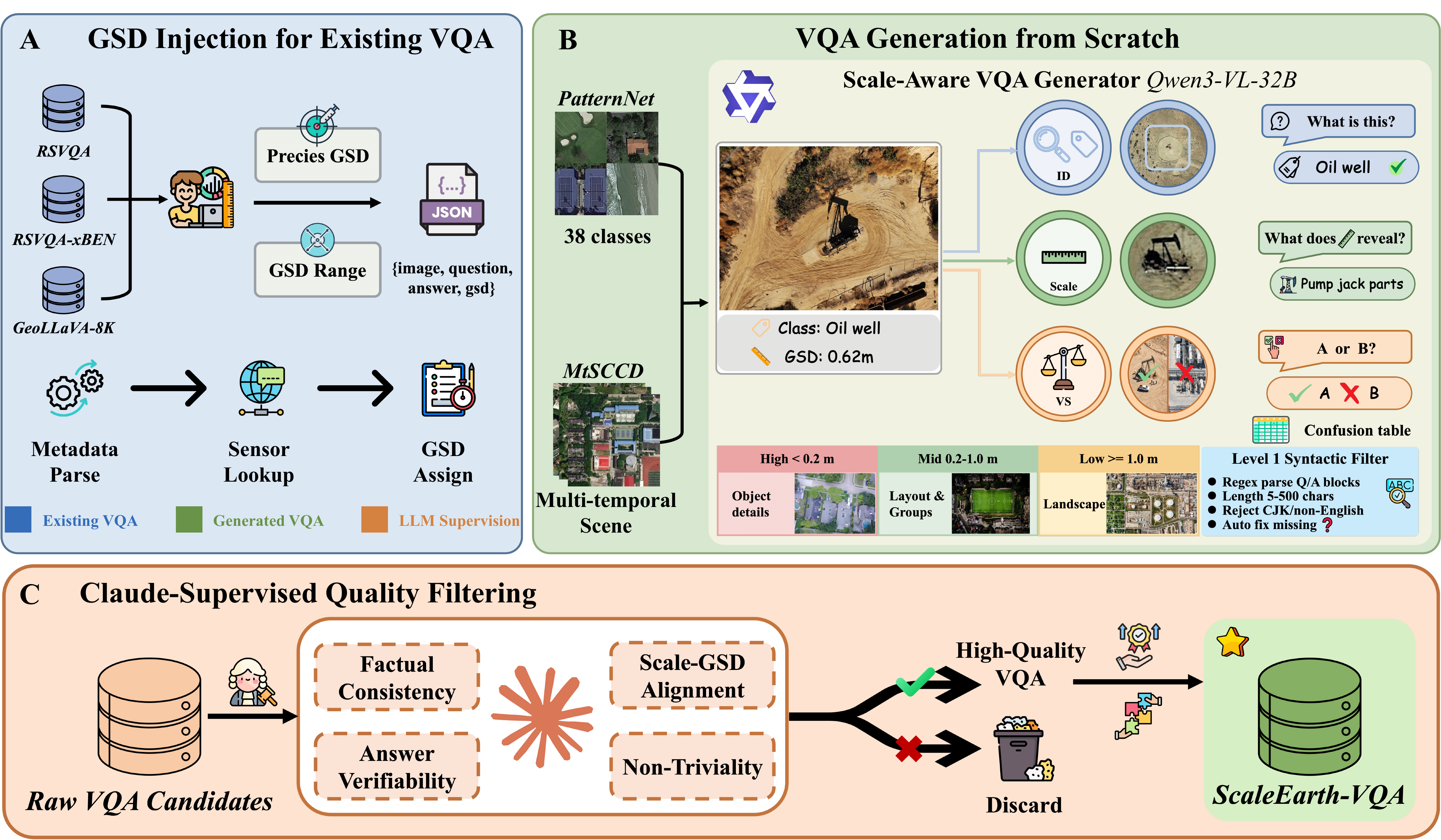}
    \caption{\textbf{GeoScale-VQA construction pipeline.}
    \textbf{(A)} GSD injection recovers per-sample GSD for existing
    corpora, through a metadata-parse, sensor-lookup, and GSD assignment resolver.
    \textbf{(B)} Scale-aware VQA generation prompts Qwen3-VL-32B with tier-conditioned instructions on PatternNet and MtSCCD, producing identification, scale-specific, and discriminative QA pairs before syntactic filtering.
    \textbf{(C)} Claude performs a four axis quality screening. Only samples satisfying all four criteria are retained in \textsc{GeoScale-VQA}.}
    \label{fig:data_pipeline}
    \vspace{-0.8em}
\end{figure}

\subsection{GSD Injection for Existing VQA}
\label{sec:dataset_gsd_injection}

The first branch, shown in Fig.~\ref{fig:data_pipeline}A, repurposes the existing RSVQA (HR/LR)~\citep{lobry2020rsvqa}, RSVQA-xBEN~\citep{lobry2021rsvqa}, and GeoLLaVA-8K~\citep{wang2025geollava} corpora. Since their GSD information is missing, implicit, or available only at the dataset level, we recover per-sample scale annotations using a three step resolver.

First, a metadata parser extracts the source sensor and acquisition tile from filenames, tile identifiers, and dataset metadata, such as Sentinel-2 MGRS tiles for xBEN and USGS HRO tiles for RSVQA-HR.
Second, a sensor lookup maps each recognized sensor to either a canonical GSD value or, for mixed-platform sources, a valid GSD range $[g_{\mathrm{lo}}, g_{\mathrm{hi}}]$. Third, GSD assignment
attaches the resolved scale to each sample as either a scalar or a range. Samples that remain unresolved are marked as unknown and are subsequently handled by the SSE-U estimator described in Sec.~\ref{sec:sseu}.

\subsection{Scale-Aware VQA Generation}
\label{sec:dataset_vqa_generation}

The GSD-injected sources alone leave the mid and low A tiers
under represented (Fig.~\ref{fig:data_landscape}(b)). We address this
imbalance by generating scale-aware VQA samples from two complementary corpora that jointly cover the missing regimes.
\textbf{PatternNet}~\citep{zhou2018patternnet} contains 38 scene
classes with GSDs from $0.062$\,m to approximately $4.7$\,m, spanning all three tiers and providing the dominant source of mid tier samples. \textbf{MtSCCD}~\citep{weixun2024mtsccd,liu2024similarity} contributes multi-temporal Chinese urban scenes at a fixed GSD of $1.01$\,m, enriching the low-tier regime with non-US urban morphology that the US-centric PatternNet cannot supply. For each image, we use \textbf{Qwen3-VL-32B-Instruct}~\citep{bai2025qwen3} to generate QA pairs. The prompt requires English answers grounded in visible evidence and forbids fabricated numbers. Its focus depends on the image's GSD tier. High-tier prompts emphasise fine-grained objects and textures. Mid-tier prompts focus on spatial layout and structural relations, while low-tier prompts target landscape patterns and land-cover semantics. For MtSCCD we further bias the generator toward land-use function and Chinese urban morphology to make the linguistic distribution match the visual distribution.

Each image yields three QA pairs. The first asks for identification, and the second targets evidence appropriate to the current GSD. The third is a discriminative question of the form ``Is this $A$ or $B$?'' Here, $B$ is a hard negative sampled from a per-class confusion table. The answer to this question must identify the correct class and justify the choice using visible evidence. Outputs that fail
syntactic checks are re-tested at a lower sampling temperature and
discarded if still unparseable.

\subsection{Claude-Supervised Quality Filtering}
\label{sec:dataset_claude_filter}

Syntactic filtering cannot reliably detect hallucinated evidence,
fabricated counts, or scale inconsistent reasoning. We therefore use
\textbf{Claude Opus 4.6} through the Anthropic API as a zero-shot judge for second stage filtering, following \citet{anthropic2025claude}
(Fig.~\ref{fig:data_pipeline}C). Each candidate QA pair is evaluated for factual consistency, answer verifiability, scale-GSD alignment, and non-triviality.
A sample is retained only when all visual claims are grounded in the visible RGB content, the answer is verifiable from the image alone, the cited evidence is consistent with the corresponding GSD tier, and the question cannot be resolved from linguistic priors alone. The accepted samples constitute the final \textsc{GeoScale-VQA} corpus used in \textbf{Stage~2} of our proposed method in Sec.~\ref{sec:method_stage2}.

\vspace{-0.8em}
\paragraph{Final corpus.}
The final corpus contains $772$K high-tier, $102$K mid-tier, $201$K low-A, and $411$K low-B QA pairs, as summarized in Fig.~\ref{fig:data_landscape}(b). Each sample is stored as a single
conversation row with image path, resolved GSD, source dataset, and QA content, requiring no changes to standard multimodal instruction-tuning data loaders.

\section{Method}
\label{sec:method}

\begin{figure}[h]
    \centering
    \includegraphics[width=\linewidth]{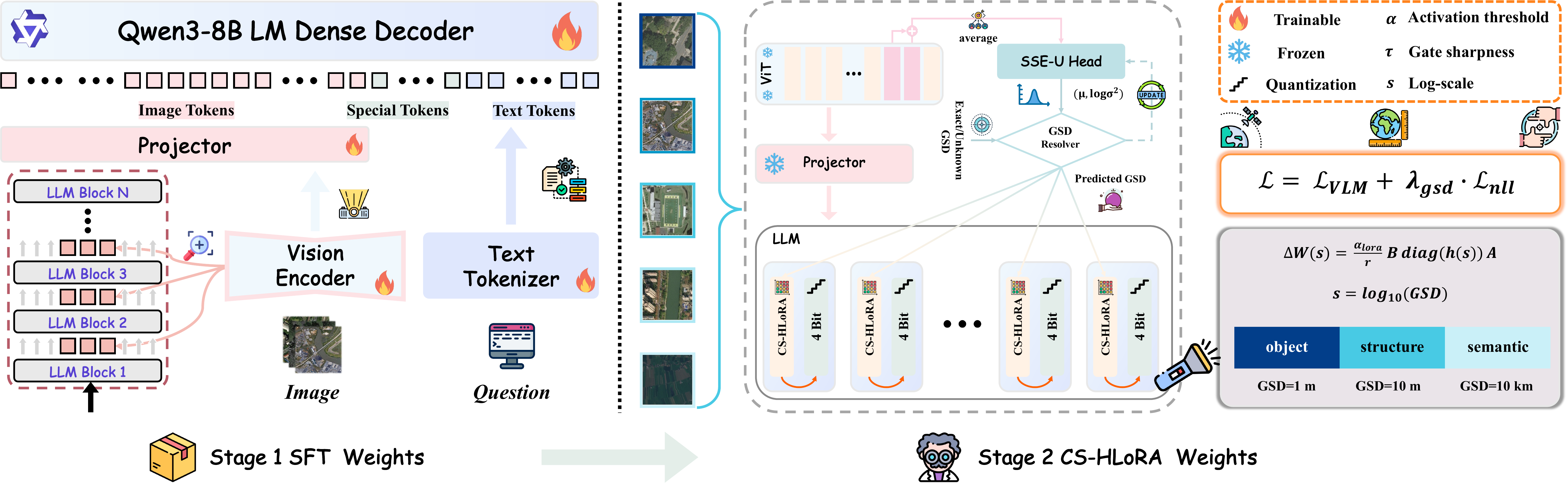}
    \caption{\textbf{ScaleEarth two stage training pipeline.}
    \textbf{Stage 1} conducts RS-domain supervised fine-tuning on the
    Qwen3-VL-8B backbone. \textbf{Stage 2} freezes the SFT-warmed backbone
    (bf16, or 4-bit NF4 for the efficiency variant) and inserts CS-HLoRA
    modules into the LLM blocks only. An SSE-U head predicts GSD and
    uncertainty $(\mu,\log\sigma^{2})$ from ViT features, and a unified
    resolver provides the effective GSD to the CS-HLoRA gate. Training
    optimizes $\mathcal{L}=\mathcal{L}_{\mathrm{VLM}}
    +\lambda_{\mathrm{gsd}}\mathcal{L}_{\mathrm{nll}}$. Special tokens and
    the multilevel ViT-to-LLM injections are Qwen3-VL's own and are left
    unchanged.}
    \label{fig:model_overview}
    \vspace{-0.8em}
\end{figure}

ScaleEarth adapts a remote sensing VLM in two stages
(Fig.~\ref{fig:model_overview}). Stage~1 performs a supervised RS-domain full parameter
 finetuning to close the natural image to the RS gap. Stage~2
freezes the warmed backbone and learns scale conditioned adaptation through
CS-HLoRA, jointly with the SSE-U scale estimation head, using the GeoScale-VQA GSD
annotations as supervision.

\subsection{Stage 1: RS-Domain Supervised Fine-Tuning}
\label{sec:method_stage1}

Qwen3-VL-8B is pretrained predominantly on natural images and inherits
systematic biases on overhead imagery, including unfamiliar viewing
geometry and a vocabulary mismatched with RS terminology. Stage~1 closes
this gap by standard supervised finetuning on
\textbf{RS-GPT4V}~\citep{xu2024rs}, a unified RS instruction corpus
covering captioning, VQA, scene understanding, and multi-turn reasoning. We
jointly update the vision encoder, projector, and language model with the
autoregressive objective
$\mathcal{L}_{\mathrm{VLM}} = -\sum_{t}\log p_{\theta}(y_t \mid y_{<t}, \mathbf{x})$
in bf16 with AdamW. No GSD conditioning is applied; the goal is an
RS-aligned backbone as a clean starting point for Stage~2.

\subsection{Stage 2: Scale-Conditioned Adaptation with CS-HLoRA and SSE-U}
\label{sec:method_stage2}

Stage~2 freezes the Stage-1 backbone and trains only two lightweight
components: \textbf{CS-HLoRA} adapters in the LLM blocks and an
\textbf{SSE-U} regression head on pooled ViT features. The frozen backbone
is kept in bf16 in our primary configuration; the efficiency variant stores
it in 4-bit NF4~\citep{dettmers2023qlora} with bf16 adapters, everything
else being identical. CS-HLoRA is applied to all attention (q/k/v/o) and FFN (gate/up/down) projections in the LLM blocks. It is not inserted into the ViT (Appendix~\ref{app:stage2}). Since Stage~1 has already adapted the vision tower to RS data, Stage~2 freezes it as a stable feature extractor. This follows common multimodal PEFT practice~\citep{liu2023visual,kuckreja2024geochat,hu2022lora}. Keeping the
ViT frozen also matters for SSE-U, which estimates GSD from the very
features the estimate then controls; gating the ViT with its own estimate
would make the features depend on the estimate itself. Joint training
couples the two modules: the gate routes adaptation by scale, and the head
learns to recover that scale from visual content alone, which is what
enables metadata free deployment (Sec.~\ref{sec:method_inference}).

\paragraph{CS-HLoRA: scale conditioned low-rank adaptation.}
A standard LoRA update $\Delta W = \frac{\alpha_{\mathrm{lora}}}{r}BA$,
with $B\in\mathbb{R}^{d\times r}$, $A\in\mathbb{R}^{r\times d}$ and rank
$r$, is shared across all inputs and thus assumes that the optimal
adaptation subspace is independent of image resolution. In remote sensing
this is restrictive, because the discriminative evidence changes
systematically with GSD. CS-HLoRA relaxes it by inserting a diagonal,
GSD-dependent gate between the two projections:
\begin{equation}
\begin{aligned}
\Delta W(s) &= \frac{\alpha_{\mathrm{lora}}}{r}\, B\,\operatorname{diag}\!\big(h(s)\big)\,A, \\
h_k(s) &= \sigma\!\left(\alpha(\tau_k-s)\right), \qquad s=\log_{10}(\mathrm{GSD}), \qquad k=0,\ldots,r-1 ,
\end{aligned}
\label{eq:cshlora_update}
\end{equation}
where $\boldsymbol{\tau}=(\tau_0,\ldots,\tau_{r-1})\in\mathbb{R}^{r}$ are
learnable per-rank activation thresholds and $\alpha>0$ is a learnable
per-projection sharpness scalar initialized to $5.0$. Each adapted
projection in every LLM block has its own $\boldsymbol{\tau}$ and
$\alpha$, all initialized with the same three-group thresholds and
$\alpha=5.0$. The gate
$h(s)\in(0,1)^r$ softly selects rank dimensions according to the input
scale; it is differentiable in $s$, so gradients flow through the gate
during joint training with SSE-U. As $\alpha\to\infty$ it reduces to hard
tier routing, while finite $\alpha$ interpolates smoothly across adjacent
GSD regimes without bucket boundaries.

\begin{figure}[h]
    \centering
    \includegraphics[width=\linewidth]{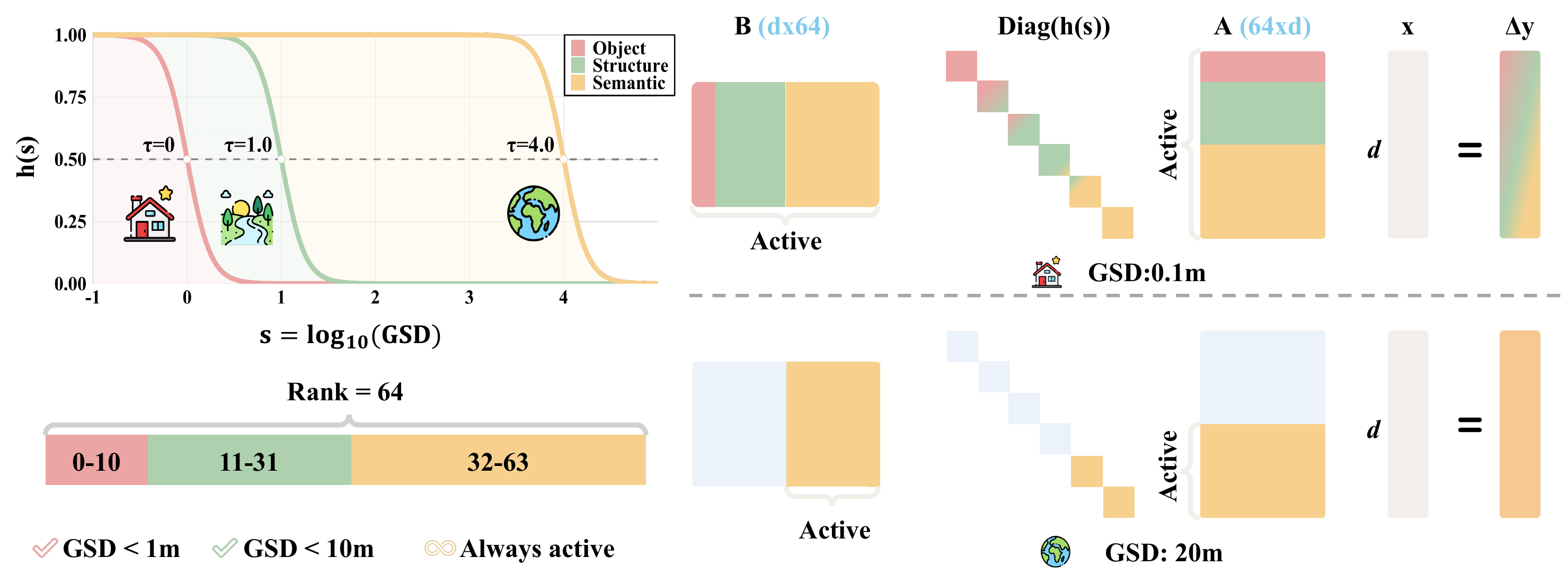}
    \vspace{-0.6em}
    \caption{\textbf{CS-HLoRA gating algorithm.}
    Each rank dimension is assigned to a scale tier through a learnable
    threshold $\tau_k$. The sigmoid gate $h_k(s)=\sigma(\alpha(\tau_k-s))$
    progressively deactivates fine-scale ranks as GSD becomes coarser:
    fine imagery (GSD $=0.1$\,m) activates object-, structure-, and
    semantic-level ranks, whereas coarse imagery (GSD $\approx20$\,m)
    primarily retains semantic ranks.}
    \label{fig:cshlora_gate}
    \vspace{-0.8em}
\end{figure}

We set $r=64$ and initialize the ranks into three physically motivated
groups following the minimum mapping unit
principle~\citep{mas2010sensitivity}, ordered as the GSD levels of
Sect.~\ref{sec:dataset}: \emph{object} ranks 0--10 with $\tau=0$ (active for
$\mathrm{GSD}\le1$\,m), \emph{structure} ranks 11--31 with $\tau=1$ (active
for $\mathrm{GSD}\le10$\,m), and \emph{semantic} ranks 32--63 with $\tau=4$
(always active). The thresholds remain trainable. The
VQA boundaries ($0.2$\,m, $1.0$\,m) and the $\tau_k$ anchors are
different kinds of quantity: the former are partition cutoff points, so each
image falls in exactly one tier, whereas the $\tau_k$ are cumulative
cutoffs in $s$; since $h_k$ decreases monotonically, group $k$ is active for
all GSD below $10^{\tau_k}$, so the groups are nested rather than disjoint,
and $\tau=4$ keeps the semantic group The
correspondence is therefore ordinal, and it holds where it should:
$\tau_{\mathrm{obj}}=0$ closes the object group at $1.0$\,m, the mid/low
boundary of GeoScale-VQA. CS-HLoRA is thus a single shared LoRA whose rank
dimensions are routed by explicit imaging scale rather than by latent,
task-agnostic mixture-of-experts signals.

\vspace{-1.0em}
\paragraph{SSE-U: heteroscedastic scale estimation.}
\label{sec:sseu}
The gate needs an image level $s$ at both training and inference; when
metadata is unavailable it must be inferred from the image. Since scale
cues vary in strength across content (urban scenes vs.\ homogeneous
vegetation or clouds), we cast this as heteroscedastic
regression~\citep{kendall2017uncertainties} and predict both the mean and
an input dependent variance of a Gaussian over log-GSD. The Qwen3-VL vision tower has no CLS token or learned class embedding. We therefore mean-pool to one image-level representation, which matches
the target: GSD is an image-level property reflected in texture, edge
density and object size across the whole image. Specifically,
$\bar{\mathbf{f}}_{\mathrm{vit}}$ averages the pre-merger patch tokens of
the third-last and second-last ViT blocks, then pools over the image's
patch tokens. SSE-U then applies a shared trunk and two linear heads:
\begin{equation}
\begin{aligned}
\mathbf{z} &= \operatorname{GELU}\!\left(W\,\operatorname{LN}(\bar{\mathbf{f}}_{\mathrm{vit}})\right), \\
\hat{\mu}(\mathbf{x}) &= W_{\mu}\mathbf{z}, \qquad
\log \hat{\sigma}^{2}(\mathbf{x}) = W_{\sigma}\mathbf{z}, \qquad
p(s\mid\mathbf{x}) = \mathcal{N}\!\left(\hat{\mu}(\mathbf{x}),\hat{\sigma}^{2}(\mathbf{x})\right).
\end{aligned}
\label{eq:sseu_dist}
\end{equation}
We clamp $\log\hat{\sigma}^{2}$ to $[-10,4]$, corresponding to $\hat{\sigma}\in[6.7\times10^{-3},\,7.4]$ in $\log_{10}$-GSD. The lower bound prevents the NLL gradient from diverging on easy samples. The upper bound remains permissive so that the head can express uncertainty on out-of-distribution inputs. SSE-U is trained with the Gaussian negative
log-likelihood over the subset $\mathcal{E}$ of samples with exact GSD,
\begin{equation}
\mathcal{L}_{\mathrm{nll}}
= \frac{1}{|\mathcal{E}|}\sum_{i\in\mathcal{E}}
\left[\frac{1}{2}\frac{\left(s_i-\hat{\mu}(\mathbf{x}_i)\right)^2}{\hat{\sigma}^{2}(\mathbf{x}_i)}
+ \frac{1}{2}\log\hat{\sigma}^{2}(\mathbf{x}_i)\right],
\label{eq:sseu_loss}
\end{equation}
whose two terms down-weight the squared error under high variance and
prevent the degenerate $\hat{\sigma}^{2}\to\infty$. The head thus learns
large uncertainty where scale cues are weak, which the inference-time
resolver uses to decide whether to trust the estimate.

\paragraph{Effective scale and joint objective.}
GSD for each GeoScale-VQA sample is exact, specified as a range, or unknown. The gate must receive a value in every case. We
define an effective scale $s_{\mathrm{eff}}$ per training step. For
exact samples, $s_{\mathrm{eff}}=\log_{10}g_{\mathrm{true}}$ with
probability $1-p_{\mathrm{e2e}}$ and $s_{\mathrm{eff}}=\hat{\mu}(\mathbf{x})$
otherwise, with $p_{\mathrm{e2e}}=0.2$, so that a fraction of updates route
through SSE-U and $\mathcal{L}_{\mathrm{VLM}}$ encourages consistency between
predicted scale and VQA performance. For range samples
$[g_{\mathrm{lo}},g_{\mathrm{hi}}]$, $s_{\mathrm{eff}}$ is drawn uniformly from
$[\log_{10}g_{\mathrm{lo}},\log_{10}g_{\mathrm{hi}}]$ at training time and
set to the log-midpoint at evaluation. For unknown samples,
$s_{\mathrm{eff}}=\hat{\mu}(\mathbf{x})$; range and unknown samples receive
$\mathcal{L}_{\mathrm{VLM}}$ gradients but not $\mathcal{L}_{\mathrm{nll}}$.
The Stage-2 objective is
\begin{equation}
\mathcal{L} \;=\; \mathcal{L}_{\mathrm{VLM}}\!\big(s_{\mathrm{eff}}\big) \;+\; \lambda_{\mathrm{gsd}}(t)\,\mathcal{L}_{\mathrm{nll}},
\label{eq:joint_loss}
\end{equation}
with $\lambda_{\mathrm{gsd}}(t)=0.3$ during the first $10\%$ of steps and
$0.1$ thereafter, stabilizing SSE-U before it equilibrates with the VQA
gradient. This three state treatment mirrors the inference time resolver
below, so deployment does not expose the model to a conditioning regime
absent from training.

\subsection{Inference: Scale-Conditioned Decoding}
\label{sec:method_inference}

Deployed RS imagery often arrives without reliable metadata. We determine the effective GSD through three mutually exclusive branches. We use caller provided metadata $g_{\mathrm{meta}}$ when available. Otherwise, we use the SSE-U estimate if it is confident ($\hat{\sigma}<\sigma_{\tau}$) and fall back to the neutral anchor $g_0$ if it is not:
\begin{equation}
g^{\star} =
\begin{cases}
g_{\mathrm{meta}}, & \text{if } g_{\mathrm{meta}}>0,\\[2pt]
10^{\hat{\mu}}, & \text{if } g_{\mathrm{meta}}\le 0 \text{ and } \hat{\sigma}<\sigma_{\tau},\\[2pt]
g_{0}, & \text{otherwise},
\end{cases}
\qquad
\hat{\sigma}=\exp\!\big(\tfrac{1}{2}\log\hat{\sigma}^{2}(\mathbf{x})\big),
\label{eq:inference_resolver}
\end{equation}
with $\sigma_{\tau}=0.3$ (a $\pm2\times$ interval in log-space) and
$g_{0}=1.0$\,m, placed at the object/structure anchor
($\tau_{\mathrm{obj}}=0$), which coincides with the mid/low boundary of
GeoScale-VQA, so that an incorrect fallback keeps the structure and semantic
groups active and degrades gracefully. The threshold is applied to the same $\hat{\sigma}$ learned for each image through Eq.~\eqref{eq:sseu_loss}, rather than a manually tuned score. Images within the training distribution yield small $\hat{\sigma}$ values, so the model uses the estimate. Images outside this distribution yield larger values and trigger the fallback. The decision to use the estimate or abstain is therefore calibrated by the model itself.

Given $g^{\star}$, the image is encoded once by the frozen ViT, SSE-U and
the CS-HLoRA gates are evaluated from those features in a single pass, and
the LLM decodes under the scale conditioned adapter. Because both run once
before the first token, they add less than $2\%$ wall-clock overhead over
base Qwen3-VL inference. For diagnostics we also support \emph{GSD
spoofing}: feeding $g^{\star}\cdot10^{\delta}$ over a log-spaced sweep of
$\delta$ (Appendix~\ref{app:spoofing}) and recording the answer at each
offset, which probes whether CS-HLoRA uses its GSD input and whether the
optimum lies at the true scale.

\section{Experiments}
\label{sec:experiments}

We evaluate whether scale-conditioned adaptation improves both fine-grained
perception and cross-scale reasoning in remote-sensing VQA. We first
describe the benchmarks and protocol, then report the main results on
XLRS-Bench and OmniEarth-Bench in both bf16 and 4-bit configurations,
robustness across GSD, and ablations that separate the mechanism from the
data and test simpler conditioning alternatives. Extended results, training
dynamics, and qualitative analyses are provided in
Appendices~\ref{app:add_results} and~\ref{appendix:qualitative}.

\subsection{Experimental Setup}
\label{sec:setup}

\noindent\textbf{Evaluation Benchmarks.}
We evaluate ScaleEarth on two complementary remote-sensing benchmarks.
\textbf{XLRS-Bench}~\citep{wang2025xlrs} targets ultra-high-resolution
imagery with sub-tasks spanning fine-grained perception and complex
reasoning. \textbf{OmniEarth-Bench}~\citep{wang2025omniearth} evaluates
cognitive ability across all six Earth-science spheres plus their
cross-sphere interactions, using observational data from 33 native sources
and 109 expert-curated tasks.

\noindent\textbf{Protocol.}
Following GeoLLaVA-8K~\citep{wang2025geollava}, we report accuracy on the
13 L-3 sub-tasks of the XLRS-Bench VQA track, grouped by their L-1
dimension (perception or reasoning), together with the unweighted macro
average. For OmniEarth-Bench we follow the original protocol and report
MCQ-style accuracy on its seven L-1 spheres. Baselines are grouped into
open-source MLLMs, closed-source MLLMs, and specialized remote-sensing
models. Unless noted otherwise, baseline numbers are taken from the
original papers or benchmark reports; every model we evaluate ourselves,
including the Qwen3-VL-8B backbone, all parity controls, and all ScaleEarth
variants, uses the same zero-shot prompting protocol through
LMMs-Eval~\citep{lmms_eval2024}, with GeoChat evaluated in its native
framework. At inference time ScaleEarth follows the scale-conditioned
decoding of Sec.~\ref{sec:method_inference}. Neither benchmark provides
per-image GSD metadata, so every ScaleEarth number in this paper is
metadata-free: each evaluation sample uses the GSD predicted by SSE-U, with
the $1.0$\,m fallback whenever $\hat\sigma>0.3$. SSE-U reaches a
$\log_{10}$-MAE of $0.12$ on held-out exact-GSD samples, its $68\%$ and
$95\%$ intervals cover $74\%$ and $96\%$ of targets, and the fallback
triggers on $8.4\%$ of benchmark samples; disabling the fallback costs
$0.7$ points. Evaluation runs on 4 A100 GPUs and the two-stage training
pipeline on 16 A100 GPUs; recipes and hyperparameters are in
Appendix~\ref{app:training-recipe}.

\subsection{Main Results}
\label{sec:main_results}

\begin{table*}[!htbp]
\centering
\caption{Results on XLRS-Bench (accuracy, \%). \textbf{Perception:} Overall/Regional Counting (OC/RC), Overall/Regional Land Use Classification (OLUC/RLUC), Object Classification (OCC), Object Color (OCL), Object Motion State (OMS), Object Spatial Relationship (OSR). \textbf{Reasoning:} Anomaly Detection (AD), Environmental Condition Reasoning (ECR), Route Planning (RP), Regional Counting with Change Detection (RCCD), Counting with Complex Reasoning (CCR). Avg.\ is the unweighted macro average over the 13 sub-tasks. Cells are ranked column-wise: \colorbox{rank1}{\strut~rose~} 1st (\textbf{bold}), \colorbox{rank2}{\strut~blue~} 2nd, \colorbox{rank3}{\strut~cream~} 3rd. The two ScaleEarth rows share architecture, data and differ only in backbone precision.}
\label{tab:xlrs_main}
\renewcommand{\arraystretch}{1.18}
\setlength{\tabcolsep}{3.2pt}
{\small
\resizebox{\textwidth}{!}{%
\begin{tabular}{l|cccccccc|ccccc|c}
\toprule
\rowcolor{headerbg}
\textbf{Method} & \multicolumn{8}{c|}{\textbf{Perception}} & \multicolumn{5}{c|}{\textbf{Reasoning}} & \textbf{Avg.} \\
\rowcolor{headerbg}
\textbf{Sub-tasks (L-3 Capability)} & OC & RC & OLUC & RLUC & OCC & OCL & OMS & OSR & AD & ECR & RP & RCCD & CCR & \\
\midrule
\rowcolor{groupbg}
\multicolumn{15}{l}{\textit{\textbf{Remote Sensing MLLMs}}} \\
GeoChat~\citep{kuckreja2024geochat}        & 16.7 & 29.0 & 2.0  & 23.0 & 21.1 & 16.8 & 35.0 & 24.2 & 33.0 & 43.0 & 10.0 & --   & 21.0 & 22.9 \\
GeoLLaVA-8K~\citep{wang2025geollava}       & 26.7 & 38.0 & \cellcolor{rank1}\textbf{49.0} & 69.0 & 41.6 & 31.6 & 65.0 & 35.0 & 67.0 & 78.0 & \cellcolor{rank1}\textbf{66.0} & \cellcolor{rank2}50.0 & \cellcolor{rank2}52.0 & 51.5 \\
GeoEyes~\citep{wang2026geoeyes}            & 38.3 & 40.0 & 24.0 & 73.5 & \cellcolor{rank1}\textbf{59.5} & \cellcolor{rank1}\textbf{66.1} & \cellcolor{rank2}68.3 & 32.2 & \cellcolor{rank2}77.0 & 80.0 & 56.0 & 40.0 & 50.0 & \cellcolor{rank3}54.2 \\
\midrule
\rowcolor{groupbg}
\multicolumn{15}{l}{\textit{\textbf{Closed-source MLLMs}}} \\
Claude 3.7 Sonnet~\citep{anthropic2025claude} & 27.6 & 22.7 & 17.4 & 68.4 & 30.5 & 29.9 & 63.6 & 27.6 & 64.8 & 78.4 & 34.5 & 27.8 & 32.6 & 40.5 \\
GPT-5.2~\citep{singh2025openai}            & 30.0 & 37.0 & 17.0 & 70.5 & 43.0 & 41.4 & \cellcolor{rank2}68.3 & 34.0 & 74.0 & 76.0 & 52.0 & 36.7 & 38.0 & 47.5 \\
\midrule
\rowcolor{groupbg}
\multicolumn{15}{l}{\textit{\textbf{Open-source MLLMs}}} \\
InternVL3-8B~\citep{zhu2025internvl3}       & 40.0 & 39.0 & 10.0 & 71.5 & 44.5 & 30.8 & 65.0 & 25.2 & \cellcolor{rank2}77.0 & \cellcolor{rank3}82.0 & 36.0 & 21.7 & 50.0 & 45.6 \\
Qwen2-VL-7B~\citep{wang2024qwen2}          & 26.7 & 40.0 & 11.0 & 73.0 & 35.9 & 34.6 & 61.7 & 31.8 & 70.0 & 81.0 & 35.0 & 46.7 & 48.0 & 45.8 \\
InternVL2.5-8B~\citep{chen2024expanding}    & 38.3 & 37.0 & 10.0 & 77.0 & 33.4 & 35.5 & 65.0 & 21.6 & 73.0 & \cellcolor{rank2}83.0 & 34.0 & \cellcolor{rank2}50.0 & 43.0 & 46.2 \\
Qwen2.5-VL-7B~\citep{bai2025qwen3}         & 33.3 & 40.0 & 31.0 & 77.0 & 40.6 & 40.5 & \cellcolor{rank3}66.7 & 36.2 & 68.0 & 72.0 & 27.0 & 38.3 & 45.0 & 47.4 \\
InternVL3-78B~\citep{zhu2025internvl3}      & 23.3 & 49.0 & 33.0 & 74.0 & 42.5 & 37.4 & \cellcolor{rank3}66.7 & 30.0 & \cellcolor{rank3}76.0 & 81.0 & 40.0 & 45.0 & 42.0 & 49.2 \\
Qwen3-VL-8B~\citep{bai2025qwen3}            & 21.7 & 50.0 & 26.0 & \cellcolor{rank2}81.5 & 46.6 & 43.1 & \cellcolor{rank3}66.7 & 30.4 & 74.0 & 79.0 & 37.0 & 43.3 & \cellcolor{rank3}51.0 & 50.0 \\
DeepEyes~\citep{zheng2025deepeyes}          & 31.7 & 41.0 & 33.0 & 75.0 & 41.6 & 38.1 & \cellcolor{rank2}68.3 & 31.4 & 70.0 & 78.0 & 46.0 & \cellcolor{rank2}50.0 & 46.0 & 50.0 \\
Qwen2.5-VL-72B~\citep{bai2025qwen3}        & 33.3 & 47.0 & 39.0 & \cellcolor{rank3}80.0 & 45.3 & 42.1 & 65.0 & 34.0 & 71.0 & 74.0 & 37.0 & 43.3 & 42.0 & 50.2 \\
Qwen3-VL-235B-A22B~\citep{bai2025qwen3}    & \cellcolor{rank1}\textbf{61.7} & \cellcolor{rank1}\textbf{82.5} & 38.6 & 36.7 & 44.0 & 39.0 & 38.9 & \cellcolor{rank1}\textbf{49.0} & 73.0 & \cellcolor{rank3}82.0 & 37.8 & 33.3 & 48.0 & 51.1 \\
\midrule
\rowcolor{ourrow}
\textbf{ScaleEarth (Ours, bf16)} &
\cellcolor{rank2}42.8 & \cellcolor{rank2}54.3 &
\cellcolor{rank2}46.0 & \cellcolor{rank1}\textbf{81.9} &
\cellcolor{rank2}57.3 & \cellcolor{rank2}55.2 &
\cellcolor{rank1}\textbf{68.7} & \cellcolor{rank2}40.9 &
\cellcolor{rank1}\textbf{77.3} & \cellcolor{rank1}\textbf{83.4} &
\cellcolor{rank2}58.8 & \cellcolor{rank1}\textbf{52.9} &
\cellcolor{rank1}\textbf{52.7} & \cellcolor{rank1}\textbf{59.4} \\
\rowcolor{ourrow}
\textbf{ScaleEarth (Ours, 4-bit)} &
\cellcolor{rank3}41.7 & \cellcolor{rank3}52.3 &
\cellcolor{rank3}44.8 & 78.6 &
\cellcolor{rank3}56.2 & \cellcolor{rank3}53.4 &
66.1 & \cellcolor{rank3}39.7 &
75.2 & 81.6 &
\cellcolor{rank3}57.3 & \cellcolor{rank3}49.1 &
50.2 & \cellcolor{rank2}57.4 \\
\bottomrule
\end{tabular}%
}
}
\end{table*}

\paragraph{Results on XLRS-Bench.}
Table~\ref{tab:xlrs_main} reports ScaleEarth in two configurations that
differ only in backbone precision: bf16 (standard LoRA, matching every
baseline) and 4-bit NF4 (QLoRA, at both training and inference). In bf16,
ScaleEarth attains the best macro average, $59.4$, against $54.2$ for
GeoEyes, the strongest RS specialist, and $51.1$ for Qwen3-VL-235B-A22B,
the strongest open generalist; the ordering holds under a sample-weighted
average as well ($56.5$ vs.\ $56.1$ for GeoEyes), so the choice of
aggregate does not change the conclusion. It is first on 6 of 13 sub-tasks
(RLUC, OMS, AD, ECR, RCCD, CCR) and second on the other seven, the only
model in the table that is first or second in every column, and it beats
GeoEyes head-to-head on 11 of 13 sub-tasks. The 4-bit configuration
absorbs the quantization penalty and still reaches $57.4$, above every
baseline, with $4.5$\,GB of base weights instead of $16$\,GB
(Appendix~\ref{app:cost}). We report the macro average because XLRS-Bench
is strongly unbalanced across sub-tasks, and all thirteen columns because
deployments care about specific tasks. Against its own backbone under the
same harness, the 4-bit model gains $+7.4$ on XLRS-Bench and $+16.4$ on
OmniEarth-Bench ($40.71$ vs.\ $24.31$).

\vspace{-1.0em}

\paragraph{Robustness across scale.}
To test whether the gains concentrate in one resolution regime, we
re-score the saved XLRS-Bench predictions of four models on three source
subsets with distinct native GSD, without retraining or re-inference
(Appendix~\ref{app:gsd_strat}). ScaleEarth is highest in every bin: $59.2$
on Toronto ($0.05$--$0.1$\,m), $57.8$ on HRSCD ($0.5$--$1$\,m) and $55.6$
on DOTA-v2.0 ($2$--$4.5$\,m). More telling is the spread. GeoLLaVA-8K is
the strongest baseline at the fine end ($58.0$) but loses $12.0$ points by
the coarse bin; GeoEyes is more even and still drops $5.6$; the Qwen3-VL-8B
backbone drops $4.8$. ScaleEarth's max--min gap is $3.6$ points, the
smallest of the four, which is the behaviour one expects from an adapter
whose active subspace follows the input scale rather than one tuned for a
dominant regime. A cross-benchmark summary of all baseline classes is given
in Appendix~\ref{app:cross_bench}.

\subsection{Ablation Study}
\label{sec:ablation}

\begin{wrapfigure}{r}{0.45\linewidth}
\vspace{-1.6em}
\centering
\includegraphics[width=0.45\textwidth]{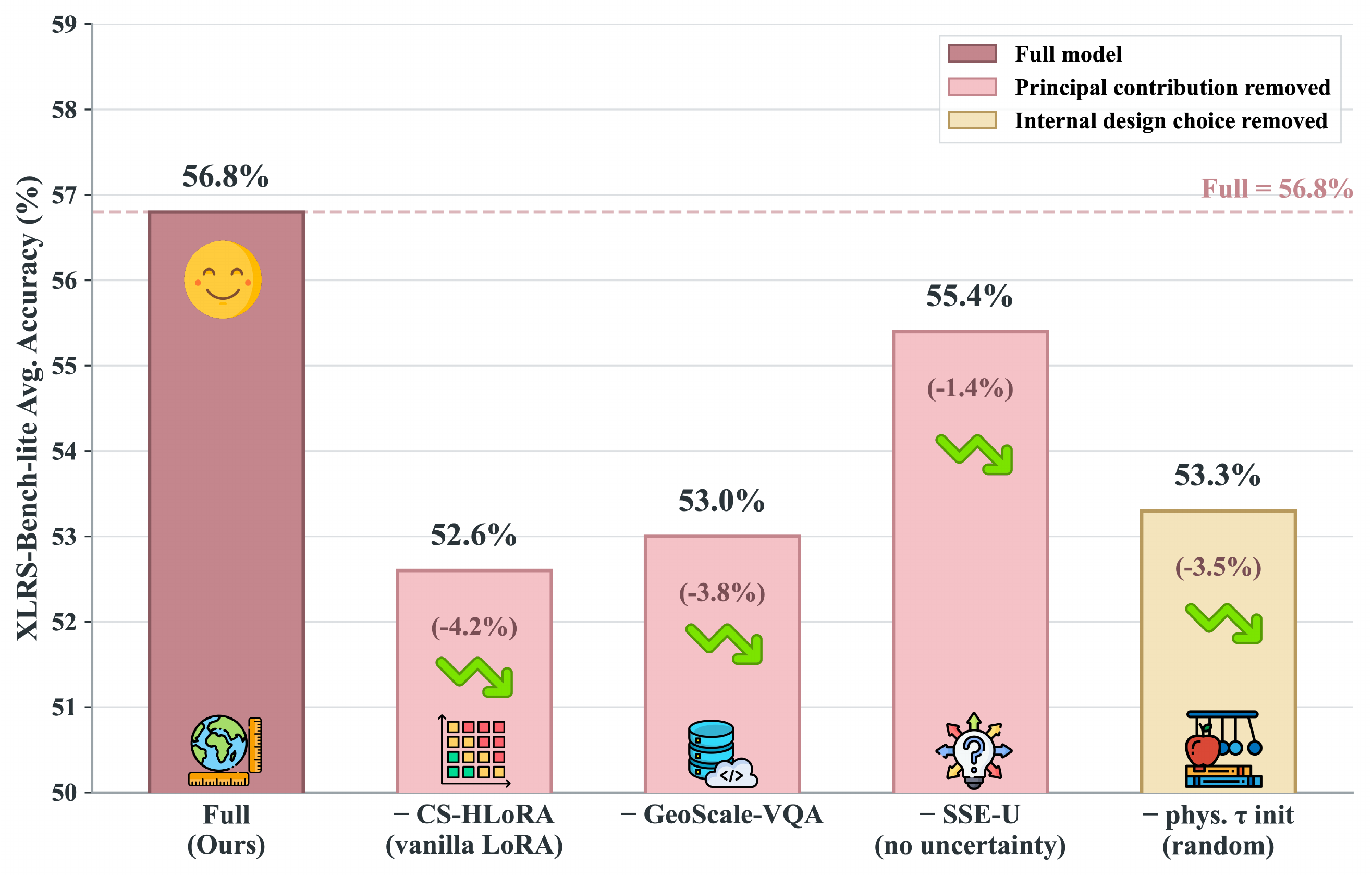}
\caption{\textbf{Leave-one-out ablation on XLRS-Bench-lite.} Bars report absolute average accuracy; in-bar values are the drop from the full model in points.}
\label{fig:ablation}
\vspace{-1.2em}
\end{wrapfigure}

\paragraph{Leave-one-out.}
We ablate each component on \textbf{XLRS-Bench-lite}, a lightweight subset
of XLRS-Bench (Fig.~\ref{fig:ablation}). Replacing CS-HLoRA with vanilla
LoRA of the same rank costs $4.2$ points, the largest single drop. Masking
the GSD metadata and scale-conditioned fields of \textsc{GeoScale-VQA},
with the samples themselves retained, costs a comparable $3.8$, showing
that the data-side scale signal is essential rather than incidental.
Randomizing the tier initialization $\tau=\{0,1,4\}$ costs $3.5$, so the
cartographic prior is actively used rather than overwritten by training.
Removing SSE-U costs $1.4$; since no benchmark sample carries GSD metadata,
this measures the value of the image-dependent estimate on every sample.
Notably, $-$SSE-U is not retrained: it fixes the gate of the trained model
at $1.0$\,m, so at inference it computes the same function class as vanilla
LoRA, yet it stays $2.8$ points ahead. GSD thus plays two distinct roles:
routing gradients to scale-specialized rank groups during training, a
specialization that persists in the learned $A$ and $B$ once the gate is
frozen, and re-weighting those groups per image at inference. Removing
SSE-U drops only the second; vanilla LoRA has neither, and the $4.2$-point
deficit decomposes accordingly into ${\sim}2.8$ from training-time routing
and ${\sim}1.4$ from inference-time gating. Re-scoring these three models
by source subset confirms that the two effects are dissociable: the
inference-time gap is smallest on HRSCD, whose native GSD straddles the
$1$\,m fallback, and larger at both ends, while the training-time gap
peaks on HRSCD (Appendix~\ref{app:ablation_ext}).

\section{Conclusion}
\label{sec:conclusion} 
ScaleEarth's closed loop among architecture, supervision, and data suggests a broader design principle for RS-VLMs. Physical variables that fundamentally distinguish overhead imagery from natural images should shape the model directly and systematically. They should not be discarded or used only as auxiliary text prompts. GSD therefore shapes both training time specialization and image dependent inference. Future work can naturally extend this principle beyond GSD to richer acquisition metadata such as spectral configuration, off-nadir angle, atmospheric conditions, and imaging geometry. It may also generalize SSE-U from aleatoric uncertainty over scale to broader uncertainty estimation over model predictions, enabling more reliable use-or-fallback decisions in high stakes applications such as disaster response and ecological monitoring.

\newpage
\bibliography{iclr2027_conference}
\bibliographystyle{iclr2027_conference}
\newpage
\appendix

\section{Additional Experiments Results}
\label{app:add_results}

\subsection{Results on OmniEarth-Bench}
\label{app:omniearth_results}

\begin{table*}[h]
\centering
\caption{Experimental results on the VQA tasks of OmniEarth-Bench.
We report accuracy for each Earth-system sphere and the arithmetic mean
across the seven spheres. ``Experts'' denotes human expert performance
evaluated on 100 examples per sphere and is included as a reference
upper bound rather than as a model baseline. Cross., Atmo., Litho.,
Ocean., Cryo., Bio., and Human. denote Cross-sphere, Atmosphere,
Lithosphere, Oceansphere, Cryosphere, Biosphere, and Human-activity
sphere, respectively. Since multiple-choice and open-ended evaluation
follow different protocols, model rankings are interpreted within each
protocol block. For the protocol-matched multiple-choice comparison, the
highest model score in each column is shown in \textbf{bold}, and the
second highest is shown with underlining. Shaded rows
highlight the main parity controls and parameter-matched adaptive PEFT
baselines used to isolate the contribution of CS-HLoRA.}
\label{tab:omniearth_main}
\renewcommand{\arraystretch}{1.18}
\setlength{\tabcolsep}{5.2pt}
\resizebox{\textwidth}{!}{%
\begin{tabular}{l|ccccccc|c}
\toprule
\rowcolor{headerbg}
\textbf{Method} &
\textbf{Cross.} &
\textbf{Atmo.} &
\textbf{Litho.} &
\textbf{Ocean.} &
\textbf{Cryo.} &
\textbf{Bio.} &
\textbf{Human.} &
\textbf{Avg.} \\
\midrule
\rowcolor{headerbg}
Experts~\citep{wang2025omniearth} &
90.00 & 96.00 & 91.00 & 95.00 & 93.00 & 97.00 & 95.00 & 93.86 \\
\midrule

\rowcolor{groupbg}
\multicolumn{9}{l}{\textit{\textbf{Multiple-choice question}}} \\
\rowcolor{groupbg}
\multicolumn{9}{l}{\textit{Closed-source MLLMs, zero-shot}} \\
Claude-3.7-Sonnet~\citep{anthropic2025claude} &
30.68 & 24.72 & 28.15 & 23.12 & 54.46 & 31.21 & 11.18 & 29.07 \\
Gemini-2.0~\citep{team2023gemini} &
16.93 & 20.83 & 38.94 & 16.94 & \underline{58.52} & 20.83 & 23.74 & 28.10 \\
GPT-4o~\citep{hurst2024gpt} &
0.04 & 9.64 & 12.80 & 13.35 & 37.48 & 1.97 & 2.76 & 11.15 \\

\midrule
\rowcolor{groupbg}
\multicolumn{9}{l}{\textit{Open-source MLLMs, zero-shot}} \\
InternVL3-72B~\citep{zhu2025internvl3} &
19.19 & 33.98 & 23.39 & 20.22 & \textbf{74.56} & 31.99 & 29.46 & 33.26 \\
InternVL3-7B~\citep{zhu2025internvl3} &
\underline{42.85} & 30.10 & 37.47 & 20.28 & 49.27 & 28.74 & 23.18 & 33.13 \\
LLaVA-OneVision-7B~\citep{li2024llava} &
19.26 & 33.69 & 28.72 & 24.54 & 46.40 & \underline{37.31} & 30.62 & 31.51 \\
InternLM-XComposer-2.5-7B~\citep{zhang2024internlm} &
19.78 & 17.45 & 28.88 & 21.06 & 40.04 & 30.67 & 24.76 & 26.09 \\
Qwen2.5-VL-7B~\citep{bai2025qwen3} &
9.85 & 9.25 & 18.65 & 13.95 & 17.85 & 10.94 & 6.23 & 12.39 \\
Qwen2.5-VL-72B~\citep{bai2025qwen3} &
3.92 & 4.82 & 22.43 & 16.27 & 5.88 & 14.91 & 8.63 & 10.98 \\
\rowcolor{backbonebg}
Qwen3-VL-8B (backbone, zero-shot)~\citep{bai2025qwen3} &
18.42 & 21.55 & 26.31 & 19.74 & 43.18 & 22.06 & 18.93 & 24.31 \\

\midrule
\rowcolor{controlbg}
\multicolumn{9}{l}{\textit{Adapted from the same Qwen3-VL-8B backbone}} \\
\rowcolor{controlbg}
Stage-1 only (RS-GPT4V SFT) &
28.74 & 26.91 & 32.40 & 24.06 & 41.55 & 28.83 & 25.62 & 29.73 \\
\rowcolor{controlbg}
Stage-1 + Standard LoRA ($r{=}64$) [B2] &
33.51 & 29.40 & 35.18 & 26.74 & 39.62 & 30.97 & 28.45 & 31.98 \\
\rowcolor{controlbg}
Stage-1 + LoRA + GSD-as-text-prompt [B3] &
34.25 & 30.18 & 36.04 & 27.30 & 39.81 & 31.55 & 29.07 & 32.60 \\

\midrule
\rowcolor{peftbg}
\multicolumn{9}{l}{\textit{Adapted from the same Stage-1 checkpoint, gated or adaptive PEFT}} \\
\rowcolor{peftbg}
Stage-1 + LoRAMoE~\citep{dou2023loramoe} &
36.71 & 31.94 & 38.45 & 28.83 & 39.94 & 33.76 & 31.18 & 34.40 \\
\rowcolor{peftbg}
Stage-1 + MoLE~\citep{wu2024mixture} &
37.85 & 32.40 & 39.21 & 29.62 & 40.07 & 34.51 & 32.30 & 35.14 \\
\rowcolor{peftbg}
Stage-1 + Bucketed MoE-LoRA ($K{=}3$, GSD-bucket-routed) [B4] &
42.13 & \underline{35.86} & \underline{41.95} & \underline{32.40} &
40.32 & 37.18 & \underline{35.42} & \underline{37.89} \\
\rowcolor{ourrow}
\textbf{ScaleEarth (Ours, full)} &
\textbf{47.20} & \textbf{38.50} & \textbf{45.30} & \textbf{34.80} &
40.80 & \textbf{39.40} & \textbf{38.95} & \textbf{40.71} \\

\midrule
\rowcolor{groupbg}
\multicolumn{9}{l}{\textit{\textbf{Open-ended question}}} \\
\rowcolor{groupbg}
\multicolumn{9}{l}{\textit{Closed-source MLLMs}} \\
Gemini-2.0~\citep{team2023gemini} &
31.48 & 38.10 & 41.67 & 24.97 & 61.49 & 27.33 & 31.85 & 36.70 \\
GPT-4o~\citep{hurst2024gpt} &
25.76 & 23.21 & 33.13 & 25.17 & 46.46 & 13.65 & 17.17 & 26.36 \\

\midrule
\rowcolor{groupbg}
\multicolumn{9}{l}{\textit{Open-source MLLMs}} \\
InternVL3-72B~\citep{zhu2025internvl3} &
29.51 & 39.14 & 27.51 & 32.45 & 53.87 & 38.29 & 34.67 & 36.49 \\
Qwen2.5-VL-72B~\citep{bai2025qwen3} &
24.78 & 22.08 & 38.62 & 31.17 & 15.23 & 20.22 & 16.87 & 24.14 \\
\bottomrule
\end{tabular}%
}
\end{table*}

\paragraph{Headline result.}
Table~\ref{tab:omniearth_main} reports per-sphere VQA accuracy on
OmniEarth-Bench. Under the protocol-matched multiple-choice setting,
ScaleEarth attains an average accuracy of $\mathbf{40.71\%}$, exceeding
the strongest zero-shot open-source baseline, InternVL3-72B
($33.26\%$), by $+7.45$ percentage points. It also outperforms the
strongest multiple-choice closed-source baseline, Claude-3.7-Sonnet
($29.07\%$), by $+11.64$ percentage points. The improvements are not
confined to the human-activity sphere, where many remote-sensing VQA
datasets are concentrated. ScaleEarth achieves the best model
performance on six of the seven spheres: Cross-sphere ($47.20\%$),
Atmosphere ($38.50\%$), Lithosphere ($45.30\%$), Oceansphere
($34.80\%$), Biosphere ($39.40\%$), and Human-activity ($38.95\%$), suggesting that the scale-aware representation learned from
GeoScale-VQA transfers beyond urban-centric remote-sensing scenes.

\paragraph{Baseline parity: separating Stage-1 SFT from CS-HLoRA.}
Several general-purpose VLMs in the upper block of
Table~\ref{tab:omniearth_main} are evaluated zero-shot, whereas
ScaleEarth first receives remote-sensing domain adaptation through
Stage~1 SFT. To separate the effect of domain SFT from the proposed
Stage~2 scale-conditioned adaptation, we include a set of parity
controls initialized from the same Qwen3-VL-8B backbone. Every
``Stage-1 + \ldots'' row starts from the same Stage-1 checkpoint, obtained
through full-parameter SFT on RS-GPT4V, and then performs Stage-2 training
on GeoScale-VQA with matched training steps, so these rows differ only in
the Stage-2 adaptation mechanism. The unmodified
backbone obtains $24.31\%$ zero-shot accuracy. Stage~1 RS-GPT4V SFT
alone increases the average to $29.73\%$, corresponding to a gain of
$+5.42$ percentage points. Adding Standard LoRA on top of Stage~1
reaches $31.98\%$, and injecting GSD as a text prompt further improves
the average to $32.60\%$. ScaleEarth reaches $40.71\%$, exceeding the
GSD-as-text-prompt control [B3] at $32.60\%$ by $+8.11$ percentage points. This
gap indicates that the gain cannot be explained solely by remote-sensing
domain SFT, additional adapter capacity, or textual exposure to GSD.

\paragraph{Comparison to gated and adaptive PEFT baselines.}
We further compare ScaleEarth with parameter-matched gated or adaptive
PEFT baselines trained from the same Stage~1 checkpoint and using the
same Stage~2 data. LoRAMoE~\citep{dou2023loramoe} and
MoLE~\citep{wu2024mixture} improve over Standard LoRA, reaching
$34.40\%$ and $35.14\%$, respectively. These results show that
input-conditioned adapter routing is beneficial, but also that generic
gated PEFT does not recover the full gain of ScaleEarth. The
GSD-bucketed MoE-LoRA baseline [B4], which hard-routes samples according
to the high-, mid-, and low-GSD partitions used in GeoScale-VQA, obtains
the strongest gated baseline average of $37.89\%$. ScaleEarth still
improves over B4 by $+2.82$ percentage points on average and by $+5.07$
percentage points on the Cross-sphere subset. This comparison suggests
that the physical routing signal is important, while the continuous
rank-wise use of that signal through CS-HLoRA provides additional
benefit over discrete bucket-level expert selection.

\paragraph{Why ScaleEarth transfers across spheres.}
We attribute the cross-sphere transfer of ScaleEarth to two factors.
First, GeoScale-VQA covers a broad range of ground sampling distances,
from sub-meter aerial imagery to $10$\,m Sentinel-2-scale observations,
thereby exposing the model to both object-centric and landscape-level
visual evidence. Second, the SSE-U module provides an uncertainty-aware
fallback when explicit GSD metadata are unavailable, which reduces the
risk of over-confident scale conditioning under metadata absence or
resolution mismatch. Together, the parity controls and the gated PEFT
comparisons indicate that ScaleEarth's improvement arises from the
combination of remote-sensing domain adaptation, physically grounded
scale conditioning, and continuous low-rank modulation.

\subsection{Cross-Benchmark Summary}
\label{app:cross_bench}

Unless stated otherwise, every ScaleEarth result in this appendix refers to
the 4-bit configuration in Table~\ref{tab:xlrs_main}. The bf16 configuration
uses the same adaptation precision as the baselines.

Table~\ref{tab:cross_bench} collects the averages from
Table~\ref{tab:xlrs_main} (XLRS-Bench macro average) and
Table~\ref{tab:omniearth_main} (OmniEarth-Bench, protocol-matched multiple
choice) for the three baseline classes. A dash means the model is not
evaluated in that table; the GeoChat OmniEarth-Bench entry was run by us
under the same protocol. On OmniEarth-Bench, ScaleEarth exceeds the strongest
closed-source baseline, Claude-3.7-Sonnet, by $+11.64$ and the strongest
open-source baseline, InternVL3-72B, by $+7.45$, and it is first on six of
the seven spheres. On XLRS-Bench, the 4-bit configuration exceeds GeoEyes by
$+3.2$, GeoLLaVA-8K by $+5.9$ and Qwen3-VL-235B-A22B by $+6.3$; the bf16
configuration widens these margins to $+5.2$, $+7.9$ and $+8.3$.

\paragraph{Cost.}
\label{app:cost}
On the same two A100 GPUs, full XLRS-Bench inference takes $2.5$ hours for
the 4-bit model, $4$ hours for the bf16 configuration and $5$ hours for the
released GeoEyes checkpoint; on a single card the 4-bit model finishes the
benchmark in $3$ hours. For deployment we merge the 4-bit adapters and serve
the model as W4A16 in vLLM with the AWQ-Marlin kernel~\citep{lin2024awq}.
Measured against a bf16 deployment on identical hardware, its LM weights
are ${\sim}3\times$ smaller, peak memory ${\sim}2.5\times$ lower, decode
throughput ${\sim}1.8\times$ higher, end-to-end latency ${\sim}1.5\times$
lower on long answers and ${\sim}1.2\times$ lower on high-resolution inputs
with short answers, and sustained concurrency ${\sim}2.5\times$ higher. On
the training side, the 4-bit backbone needs $4.5$\,GB of base weights
instead of $16$\,GB and $30$\,GB of training memory instead of $150$\,GB.
At 8K input resolution, ScaleEarth uses $68\%$ less peak GPU memory than
GeoLLaVA-8K.

\begin{table}[h]
\centering
\small
\renewcommand{\arraystretch}{1.15}
\setlength{\tabcolsep}{6pt}
\caption{Average accuracy of the three baseline classes and ScaleEarth on
both benchmarks. XLRS-Bench values are the macro averages of the available sub-task scores in
Table~\ref{tab:xlrs_main}. GeoChat has no reported RCCD score. OmniEarth-Bench is the protocol-matched
multiple-choice average of Table~\ref{tab:omniearth_main}. A dash means the
model is not evaluated in that table.}
\label{tab:cross_bench}
\resizebox{\linewidth}{!}{%
\begin{tabular}{llccc}
\toprule
\rowcolor{headerbg}
\textbf{Class} & \textbf{Model} & \textbf{Params} & \textbf{XLRS-Bench Avg.} & \textbf{OmniEarth Avg.\ (MCQ)} \\
\midrule
\multirow{4}{*}{Closed-source generalist}
 & GPT-4o             & --   & --   & 11.15 \\
 & Gemini-2.0         & --   & --   & 28.10 \\
 & Claude-3.7-Sonnet  & --   & 40.5 & 29.07 \\
 & GPT-5.2            & --   & 47.5 & --    \\
\midrule
\multirow{4}{*}{Open-source generalist}
 & Qwen2.5-VL-72B             & 72B  & 50.2 & 10.98 \\
 & Qwen3-VL-8B (our backbone) & 8B   & 50.0 & 24.31 \\
 & InternVL3-72B              & 72B  & --   & 33.26 \\
 & Qwen3-VL-235B-A22B         & 235B & 51.1 & --    \\
\midrule
\multirow{3}{*}{RS specialist}
 & GeoChat      & 7B & 22.9 & 20.5 \\
 & GeoLLaVA-8K  & 7B & 51.5 & --   \\
 & GeoEyes      & 7B & 54.2 & --   \\
\midrule
\rowcolor{ourrow}
\textbf{Ours} & \textbf{ScaleEarth (bf16)}  & \textbf{8B}        & \textbf{59.4} & --             \\
\rowcolor{ourrow}
\textbf{Ours} & \textbf{ScaleEarth (4-bit)} & \textbf{8B, 4-bit} & \textbf{57.4} & \textbf{40.71} \\
\bottomrule
\end{tabular}}

\end{table}

\subsection{Results Stratified by GSD}
\label{app:gsd_strat}

We split the XLRS-Bench evaluation samples into three GSD bins and re-score
the existing predictions of four models in each bin, with no retraining and
no re-inference. Each bin comes from a different source subset of
XLRS-Bench, so the bins are distinct sensors and resolutions rather than
resampled copies of the same imagery: Toronto with a native GSD of
$0.05$--$0.1$\,m, HRSCD with $0.5$--$1$\,m, and DOTA-v2.0 with
$2$--$4.5$\,m~\citep{xia2018dota,ding2021object}.

\begin{table}[h]
\centering
\small
\renewcommand{\arraystretch}{1.15}
\setlength{\tabcolsep}{6pt}
\caption{XLRS-Bench accuracy stratified by native GSD. Predictions are
re-scored by source subset; no model is retrained or re-run. The last row
is the max--min spread across the three bins (lower is more stable).}
\label{tab:gsd_strat}
\resizebox{\linewidth}{!}{%
\begin{tabular}{lrcccc}
\toprule
\rowcolor{headerbg}
\textbf{GSD bin (source)} & $N$ & \textbf{Qwen3-VL-8B} & \textbf{GeoLLaVA-8K} & \textbf{GeoEyes} & \textbf{ScaleEarth} \\
\midrule
$<0.5$\,m (Toronto)          & 1{,}240 & 46.6 & 58.0 & 56.5 & \textbf{59.2} \\
$0.5$--$2$\,m (HRSCD)        & 1{,}680 & 51.4 & 51.5 & 55.6 & \textbf{57.8} \\
$2$--$10$\,m (DOTA-v2.0)     & 1{,}510 & 51.3 & 46.0 & 50.9 & \textbf{55.6} \\
\midrule
Average (all GSD)            & 4{,}430 & 50.0 & 51.5 & 54.2 & \textbf{57.4} \\
Max--min gap across bins     &         & 4.8  & 12.0 & 5.6  & \textbf{3.6}  \\
\bottomrule
\end{tabular}}

\end{table}

Qwen3-VL-8B drops most below $0.5$\,m, where fine object detail carries the
answer. GeoLLaVA-8K is the strongest baseline in that bin, which matches its
design for 8K inputs, but it loses $12.0$ points by the $2$--$10$\,m bin.
GeoEyes is more even across bins and still drops $5.6$ points at the coarse
end. ScaleEarth is highest in all three bins, and its max--min gap of $3.6$
points, against $4.8$--$12.0$ for the baselines, makes it the most stable of
the four across scales.

\subsection{Ablation Protocol and GSD-Stratified Ablations}
\label{app:ablation_ext}

\paragraph{Protocol.}
Except for the inference-only removal of SSE-U, all configurations in
Fig.~\ref{fig:ablation} undergo Stage-2 training from the same Stage-1
checkpoint obtained through full-parameter SFT on RS-GPT4V. They use the
same GeoScale-VQA samples and matched training steps, with the changes
specified below.
\emph{$-$GeoScale-VQA} is not a corpus replacement: the same samples are
retained, while their GSD metadata and scale-conditioned fields are masked,
so there are no tier-specific prompts and no scale-aware supervision. This
isolates the data-side scale signal while keeping the training samples and
the architecture unchanged. \emph{$-$SSE-U} is the one row that is not retrained: it loads the fully
trained CS-HLoRA weights and removes the entire SSE-U module at inference,
including both the GSD prediction and the uncertainty estimation, so no
image-dependent scale is inferred at test time. Since neither benchmark
provides per-image GSD, every sample then uses the fixed fallback of
$1.0$\,m, which yields the same CS-HLoRA gate vector for all images. The
$-$SSE-U model and vanilla LoRA therefore have equivalent forward
computation at inference and differ only in how their parameters were
learned.
\emph{Random $\tau$} samples all $64$ thresholds independently from
$\mathcal{N}(0,1)$, so the physically ordered object/structure/semantic
grouping is not preserved; the $\alpha$ initialization, rank, optimizer and
schedule are unchanged.

\paragraph{Two roles of GSD.}
GSD plays two separable roles in CS-HLoRA. (1) \emph{Training-time
routing}: the gate sends gradients to the object, structure and semantic
rank groups according to scale, so each group specializes to its own
resolution band. A single vanilla-LoRA subspace instead has to fit all
scales at once and absorbs cross-scale gradient interference. This benefit
is baked into the learned $B$ and $A$ and survives even when the gate is
later frozen. (2) \emph{Inference-time gating}: at test time the gate
re-weights the already-specialized groups per image according to the
predicted GSD. Removing SSE-U keeps (1) and drops (2); vanilla LoRA has
neither. The $4.2$-point deficit in Fig.~\ref{fig:ablation} therefore
splits into ${\sim}2.8$ points from training-time specialization and
${\sim}1.4$ points from inference-time gating. That $-$SSE-U suffers less
than vanilla LoRA shows that the dominant contribution of GSD is (1): it
structures the parameter space during training rather than acting as a
runtime crutch. Both FiLM and CS-HLoRA respond continuously to GSD at
inference, but CS-HLoRA achieves $40.71$ on OmniEarth-Bench compared with
$36.28$ for FiLM, a difference of $4.43$ points
(Table~\ref{tab:conditioning}). In CS-HLoRA, GSD also guides scale-dependent
training, and the resulting specialization remains in the learned
parameters when the inference gate is fixed.

\paragraph{Ablations by GSD bin.}
A natural objection is that the $2.8$-point separation between $-$SSE-U and
vanilla LoRA could be a bias toward the $1$\,m fallback, or variance, rather
than a property of scale-conditioned training. Table~\ref{tab:ablation_gsd}
re-scores the saved predictions of the three models on the three source
subsets of Table~\ref{tab:gsd_strat}, again without retraining or
re-inference. The two difference columns move in opposite directions with
scale, which rules out both alternatives. The inference-time channel
(Full $-$ ($-$SSE-U)) is smallest at HRSCD ($0.6$), whose native GSD
straddles the fixed $1$\,m fallback, and larger at both ends ($2.4$ and
$1.8$): gating matters where the fallback is wrong. The training-time gap
(($-$SSE-U) $-$ vanilla) does the reverse and peaks at HRSCD ($3.4$): near
the fallback, $-$SSE-U runs at its correct scale while vanilla LoRA still
pays the full cost of having no scale-specialized subspace. A $1$\,m bias
would flatten both columns rather than produce opposite scale dependencies,
and the total Full $-$ vanilla gap is nearly constant across bins
($4.3$, $4.0$, $4.3$), matching the $4.2$-point aggregate of
Fig.~\ref{fig:ablation}. Both effects are therefore genuine and dissociable.

\begin{table}[h]
\centering
\small
\renewcommand{\arraystretch}{1.15}
\setlength{\tabcolsep}{5pt}
\caption{Leave-one-out ablations re-scored by source subset (no retraining,
no re-inference). Differences are computed from the adjacent columns.}
\label{tab:ablation_gsd}
\resizebox{\linewidth}{!}{%
\begin{tabular}{lccccc}
\toprule
\rowcolor{headerbg}
\textbf{Source (native GSD)} & \textbf{Full} & $-$\textbf{SSE-U} & \textbf{Full} $-$ ($-$\textbf{SSE-U}) & \textbf{vanilla LoRA} & ($-$\textbf{SSE-U}) $-$ \textbf{vanilla} \\
\midrule
Toronto ($0.05$--$0.1$\,m)   & 59.2 & 56.8 & $+2.4$ & 54.9 & $+1.9$ \\
HRSCD ($0.5$--$1$\,m)        & 57.8 & 57.2 & $+0.6$ & 53.8 & $+3.4$ \\
DOTA-v2.0 ($2$--$4.5$\,m)    & 55.6 & 53.8 & $+1.8$ & 51.3 & $+2.5$ \\
\bottomrule
\end{tabular}}

\end{table}

\subsection{Data or Continuous Scale Conditioning?}
\label{app:factorial}

\paragraph{Factorial design.}
The factorial varies the corpus and the conditioning mechanism
independently. Factor A is the corpus. A0 is FIT-RS~\citep{luo2024skysensegpt},
an external instruction corpus built on the STAR imagery
collection~\citep{li2025star}. It shares no images and no QA pairs with
GeoScale-VQA, and none of our annotation, generation or filtering pipeline
was applied to it. Its GSD comes from the resolution metadata published with
FIT-RS, which spans $0.15$--$4$\,m. A1 is GeoScale-VQA. Factor B is either
standard LoRA at rank $64$ with no GSD input (B0) or CS-HLoRA (B1). All four
cells share the Stage-1 checkpoint, sample count, schedule,
trainable-parameter budget and evaluation harness, and we report the mean
over three seeds. FIT-RS supplies $1.42$M training samples against $1.49$M
for GeoScale-VQA. We randomly subsample each corpus to $N{=}300$K in all
four cells so that the twelve runs fit into the available compute. The
reduction applies equally to every cell, all conclusions are read from
differences at the same $N$, and the published full-corpus result is given
as a reference row.

\begin{table}[h]
\centering
\small
\renewcommand{\arraystretch}{1.15}
\setlength{\tabcolsep}{5pt}
\caption{Two-factor design over corpus (A) and conditioning mechanism (B),
with the corresponding main-effect decomposition. All cells use the same
Stage-1 checkpoint, $N{=}300$K Stage-2 samples, matched steps and
parameter budget; means over three seeds. The reference and shuffled rows
use the full corpus.}
\label{tab:factorial}
\resizebox{\linewidth}{!}{%
\begin{tabular}{lllcc}
\toprule
\rowcolor{headerbg}
\textbf{Cell} & \textbf{Corpus (A)} & \textbf{Conditioning (B)} & \textbf{OmniEarth Avg.} & \textbf{XLRS macro-avg} \\
\midrule
A0B0 & FIT-RS, external, $N{=}300$K & standard LoRA $r{=}64$ & 29.0 & 50.9 \\
A0B1 & FIT-RS, external, $N{=}300$K & CS-HLoRA               & 34.5 & 53.9 \\
A1B0 & GeoScale-VQA, $N{=}300$K     & standard LoRA $r{=}64$ & 31.8 & 52.4 \\
\rowcolor{ourrow}
\textbf{A1B1} & \textbf{GeoScale-VQA}, $N{=}300$K & \textbf{CS-HLoRA} & \textbf{38.0} & \textbf{55.8} \\
\midrule
reference only & GeoScale-VQA, full $N{=}1.49$M        & CS-HLoRA & 40.71 & 57.4 \\
shuffled GSD   & GeoScale-VQA, full, GSD permuted      & CS-HLoRA & 31.4  & 52.8 \\
\midrule
\rowcolor{headerbg}
\multicolumn{2}{l}{\textbf{Decomposition}} & \textbf{Definition} & \textbf{OmniEarth} & \textbf{XLRS} \\
\multicolumn{2}{l}{Data main effect}      & A1B0 $-$ A0B0                              & $+2.8$          & $+1.5$          \\
\multicolumn{2}{l}{Algorithm main effect} & A0B1 $-$ A0B0                              & $\mathbf{+5.5}$ & $\mathbf{+3.0}$ \\
\multicolumn{2}{l}{Interaction}           & (A1B1 $-$ A1B0) $-$ (A0B1 $-$ A0B0)        & $+0.7$          & $+0.4$          \\
\multicolumn{2}{l}{Total}                 & A1B1 $-$ A0B0                              & $+9.0$          & $+4.9$          \\
\bottomrule
\end{tabular}}

\end{table}

The primary comparison is A0B1 against A1B0. At an identical sample count of
$300$K, CS-HLoRA trained on FIT-RS reaches $34.5$ on OmniEarth-Bench, whereas
standard LoRA trained on GeoScale-VQA reaches $31.8$, even though FIT-RS is
external and shares no imagery with GeoScale-VQA. A0B1 also exceeds the
$31.98$ that standard LoRA reaches on the full $1.49$M corpus
(Table~\ref{tab:omniearth_main}), which gives that baseline about five
times more data. The three effects decompose a total gain of $9.0$ points on
OmniEarth-Bench: $5.5$ from the algorithm main effect, $2.8$ from the data
main effect and $0.7$ from the interaction. The XLRS-Bench effects follow the
same ordering, with an algorithm main effect of $3.0$ out of a $4.9$-point
total; XLRS-Bench spans a narrower GSD range than OmniEarth-Bench, so a
smaller absolute effect is expected.

\paragraph{GSD-shuffle control.}
We keep the full GeoScale-VQA corpus, the CS-HLoRA architecture, the
parameter count and the training budget unchanged, and randomly permute the
GSD label across the training set, so only the image-to-scale correspondence
changes. The shuffled model falls back to the level of the unconditioned
baseline ($31.4$ on OmniEarth-Bench, $52.8$ on XLRS-Bench). Since the corpus
is identical in both runs, the gain comes from the scale variable and not
from corpus size or quality.

\paragraph{Simpler conditioning.}
We add two simpler baselines: an auxiliary GSD-prediction head without
gating, and a FiLM module~\citep{perez2018film} conditioned on
$\log$-GSD. All rows of Table~\ref{tab:conditioning} share the Stage-1
checkpoint, the GeoScale-VQA Stage-2 corpus, matched steps and comparable
trainable parameters. The last column gives the response when only the
supplied GSD is swept, with image and question fixed
(Appendix~\ref{app:spoofing}).

\begin{table}[h]
\centering
\small
\renewcommand{\arraystretch}{1.15}
\setlength{\tabcolsep}{6pt}
\caption{Conditioning strategies on OmniEarth-Bench and their response to
the GSD-spoofing sweep of Appendix~\ref{app:spoofing}. All rows share the
Stage-1 checkpoint and the Stage-2 corpus.}
\label{tab:conditioning}
\resizebox{\linewidth}{!}{%
\begin{tabular}{lcl}
\toprule
\rowcolor{headerbg}
\textbf{Conditioning strategy} & \textbf{OmniEarth Avg.} & \textbf{Response to GSD sweep} \\
\midrule
Standard LoRA, no GSD [B2]                 & 31.98 & Flat, 74--76\% \\
GSD as text prompt [B3]                    & 32.60 & Flat, 76--78\% \\
Auxiliary GSD-prediction head, no gating   & 32.55 & Flat, 73--77\% \\
FiLM on $\log$-GSD, parameter-matched      & 36.28 & Bell-shaped, ${\sim}79.3$\% at true GSD \\
Bucketed MoE-LoRA, GSD-routed [B4]         & 37.89 & Stepwise, ${\sim}15$-point drop from $0.18$\,m to $0.22$\,m \\
\rowcolor{ourrow}
\textbf{CS-HLoRA (ours)}                   & \textbf{40.71} & Smooth bell, ${\sim}86$\% at true GSD \\
\bottomrule
\end{tabular}}

\end{table}

SSE-U shows that GSD is decodable from frozen features, but decodability
does not mean the computation uses it. B2 and the auxiliary head are trained
on the same scale-annotated corpus; the auxiliary head predicts GSD
accurately, yet its answers do not change from $0.01$\,m to $30$\,m.
Supervision alone does not create a path through which scale can reach the
update. FiLM does condition continuously, improves the average to $36.28$
and gives a bell-shaped response, so continuity is a real part of the
effect, but it remains $4.43$ points below CS-HLoRA. A feature-wise affine
transform rescales activations uniformly, whereas CS-HLoRA selects
\emph{which} low-rank directions are active at a given scale, the per-rank
specialization of Fig.~\ref{fig:tau}. Gating inside the adapter subspace
also costs only $r{=}64$ values per layer rather than an $O(d)$
$\gamma$/$\beta$ generator, and it admits a physical initialization. B2
reaches $31.98$ against $29.73$ for Stage~1 alone, so at most $2.25$ of the
$10.98$-point Stage-2 gain comes from added data and capacity; the remainder
requires CS-HLoRA.

\subsection{Sensitivity of the Scale Priors}
\label{app:prior_sens}

We varied one prior at a time and held everything else at the default,
using the same Stage-1 checkpoint, corpus, schedule and parameter budget in
every run. Both $\tau$ and $\alpha$ are trained in Stage~2, so these values
are an initialization rather than tuned hyperparameters, and
Fig.~\ref{fig:stage2-overview}(d) shows the gates converge to the intended
partition without supervision. Three seeds at the default give standard
deviations of $0.35$ on XLRS-Bench-lite and $0.30$ on OmniEarth-Bench.

\begin{table}[h]
\centering
\small
\renewcommand{\arraystretch}{1.15}
\setlength{\tabcolsep}{6pt}
\caption{One-at-a-time sensitivity of rank allocation, gate anchors, and the
number of groups. Default rows are repeated for reference.}
\label{tab:prior_axes}
\resizebox{\linewidth}{!}{%
\begin{tabular}{llcc}
\toprule
\rowcolor{headerbg}
\textbf{Axis} & \textbf{Variant} & \textbf{XLRS-Bench-lite} & \textbf{OmniEarth} \\
\midrule
\multirow{3}{*}{A. Rank allocation}
 & $0$--$10$ / $11$--$31$ / $32$--$63$ (default) & \textbf{58.0} & \textbf{40.71} \\
 & inverted $32$ / $21$ / $11$                    & 56.4 & 39.20 \\
 & extreme $2$ / $2$ / $60$                       & 55.8 & 38.60 \\
\midrule
\multirow{5}{*}{B. Gate anchors / $K$}
 & gate cutoffs $1$ / $10$ / $10^{4}$\,m, $K{=}3$ (default) & \textbf{58.0} & \textbf{40.71} \\
 & shifted $+0.5$ in $\log_{10}\mathrm{GSD}$            & 57.3 & 40.10 \\
 & shifted $-0.5$ in $\log_{10}\mathrm{GSD}$            & 57.4 & 40.20 \\
 & $K{=}2$                                              & 56.6 & 39.30 \\
 & $K{=}4$                                              & 57.7 & 40.40 \\
\bottomrule
\end{tabular}}

\end{table}

Both axes move accuracy by at most $2.2$ points. Rank allocation spans
$2.11$ points on OmniEarth-Bench: capacity is fixed at $r{=}64$, and
starving the object group costs accuracy on the tier holding $52\%$ of the
corpus. Varying the gate anchors and group count spans $1.41$ points on
OmniEarth-Bench, with the largest drop occurring at $K{=}2$. Shifting the
anchors by $\pm0.5$ in $\log_{10}\mathrm{GSD}$ costs at most $0.61$ points. The worst
variant reaches $38.60$, above the parameter-matched baseline B4 at $37.89$.

\begin{table}[h]
\centering
\small
\renewcommand{\arraystretch}{1.15}
\setlength{\tabcolsep}{8pt}
\caption{Joint sweep of rank allocation and a global shift $\Delta\tau$ of
the tier anchors on XLRS-Bench-lite under a matched, reduced Stage-2
training budget. The default configuration is rerun under the same reduced
budget.}
\label{tab:prior_grid}
\begin{tabular}{lccc}
\toprule
\rowcolor{headerbg}
\textbf{Rank allocation (obj / str / sem)} & $\Delta\tau{=}-0.5$ & $\Delta\tau{=}0$ & $\Delta\tau{=}+0.5$ \\
\midrule
$11$ / $21$ / $32$ (default)  & 57.4 & \textbf{58.0} & 57.3 \\
$21$ / $21$ / $22$ (uniform)  & 56.8 & 57.2 & 56.6 \\
$32$ / $21$ / $11$ (inverted) & 55.9 & 56.4 & 55.7 \\
\bottomrule
\end{tabular}
\end{table}

One-dimensional controls at the default setting: $\alpha$ initialization
$\{2.5, 5.0, 10.0\}$ gives $57.6$ / $\mathbf{58.0}$ / $57.7$, and tier
boundaries $\{0.15$--$0.8,\ \mathbf{0.2}$--$\mathbf{1.0},\ 0.3$--$1.5\}$\,m
give $57.7$ / $\mathbf{58.0}$ / $57.6$. Both are within noise, consistent
with the near-uniform learned $\alpha$ in Fig.~\ref{fig:stage2-overview}.
The default and uniform rank allocations in Table~\ref{tab:prior_grid}
remain within a $1.4$-point band across the tested shifts. At
$\Delta\tau=0$, the inverted and uniform allocations reduce accuracy by
$1.6$ and $0.8$ points, respectively. Randomizing $\tau$ gives $54.5$, a
drop of $3.5$ points from the matched default of $58.0$.

\begin{table}[h]
\centering
\small
\renewcommand{\arraystretch}{1.15}
\setlength{\tabcolsep}{8pt}
\caption{Learned group means of $\tau$ after Stage~2 for different
initializations, with weight decay anchored to each configuration's own
initialization so that any drift toward the default values is
data-driven. The last column is the peak of the GSD-spoofing curve
(Appendix~\ref{app:spoofing}) at a true GSD of $0.15$\,m.}
\label{tab:prior_learned}
\begin{tabular}{lll}
\toprule
\rowcolor{headerbg}
\textbf{Initial $\tau$ (obj / str / sem)} & \textbf{Learned mean} & \textbf{Spoofing peak (true $0.15$\,m)} \\
\midrule
$-0.5$ / $0.5$ / $3.5$                & $-0.28$ / $0.67$ / $3.58$ & $0.15$\,m \\
$\mathbf{0.0}$ / $\mathbf{1.0}$ / $\mathbf{4.0}$ (default) & $0.07$ / $1.09$ / $4.02$ & $0.15$\,m \\
$+0.5$ / $1.5$ / $4.5$                & $0.34$ / $1.38$ / $4.47$  & $0.17$\,m \\
Random $\mathcal{N}(0,1)$             & ordering not recovered    & $0.41$\,m (mis-peaked) \\
\bottomrule
\end{tabular}
\end{table}

The negatively and positively shifted initializations produce peaks at
$0.15$ and $0.17$\,m, respectively, both close to the true GSD of $0.15$\,m.
Random initialization does not recover the intended ordering and produces
a peak at $0.41$\,m.
The physical initialization therefore supplies the gate topology, not tuned
constants, and $\tau$ and $\alpha$ remain learnable throughout. Randomizing
$\tau$ costs $3.5$ points (Fig.~\ref{fig:ablation}), yet with the physical
initialization the learned gates converge to the object/structure/semantic
partition without supervision on gate values
(Fig.~\ref{fig:stage2-overview}(d)), and the resulting groups are
functionally distinct: ranks $0$--$10$ predict texture, $11$--$31$ edge
geometry and $32$--$63$ scene identity (Fig.~\ref{fig:tau}). Gate sharpness
enters only through $\sigma(\alpha(\tau_k - s))$, where $\alpha\to\infty$
gives hard tier routing and a finite $\alpha$ sets the transition width near
a cutoff (Sec.~\ref{sec:method_stage2}). Since $\alpha$ is trained per
layer, its initialization sets that width rather than fixing it.

\subsection{Training Dynamics}
\label{app:dynamics}

\begin{figure*}[t]
\centering
\includegraphics[width=\textwidth]{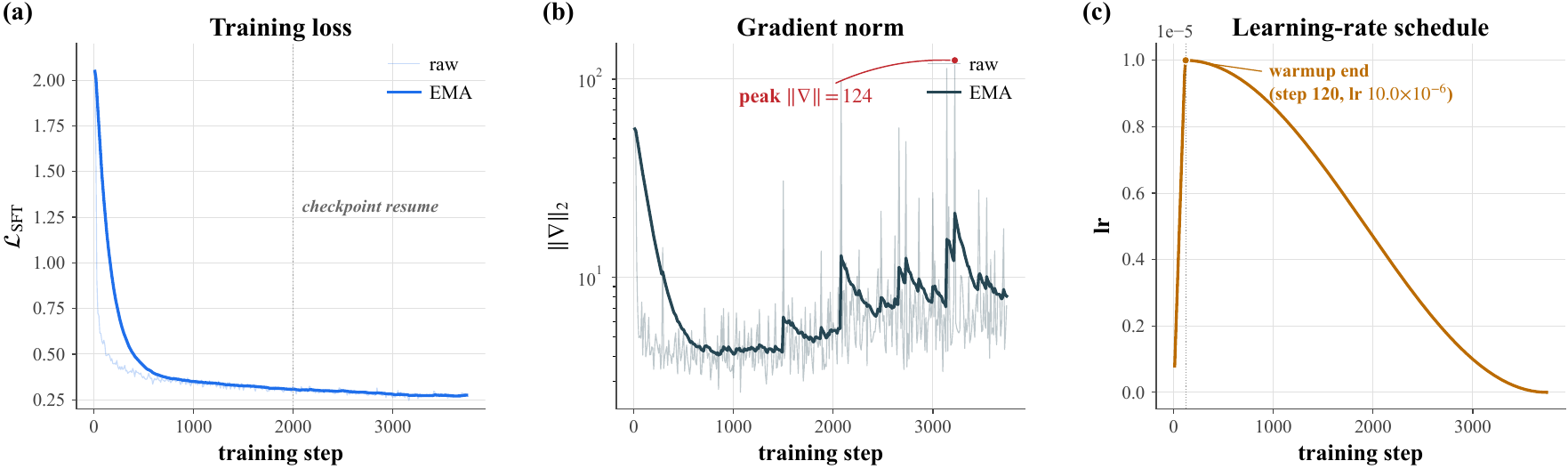}
\caption{\textbf{Stage~1 full-parameter RS SFT training dynamics.}
(a)~SFT cross-entropy loss $\mathcal{L}_{\text{SFT}}$ with raw and EMA
curves. (b)~Per-step gradient norm $\|\nabla\|_2$ on a logarithmic scale.
(c)~Cosine learning-rate schedule with linear warmup.}
\label{fig:stage1-overview}
\end{figure*}

\begin{figure*}[!ht]
\centering
\includegraphics[width=\textwidth]{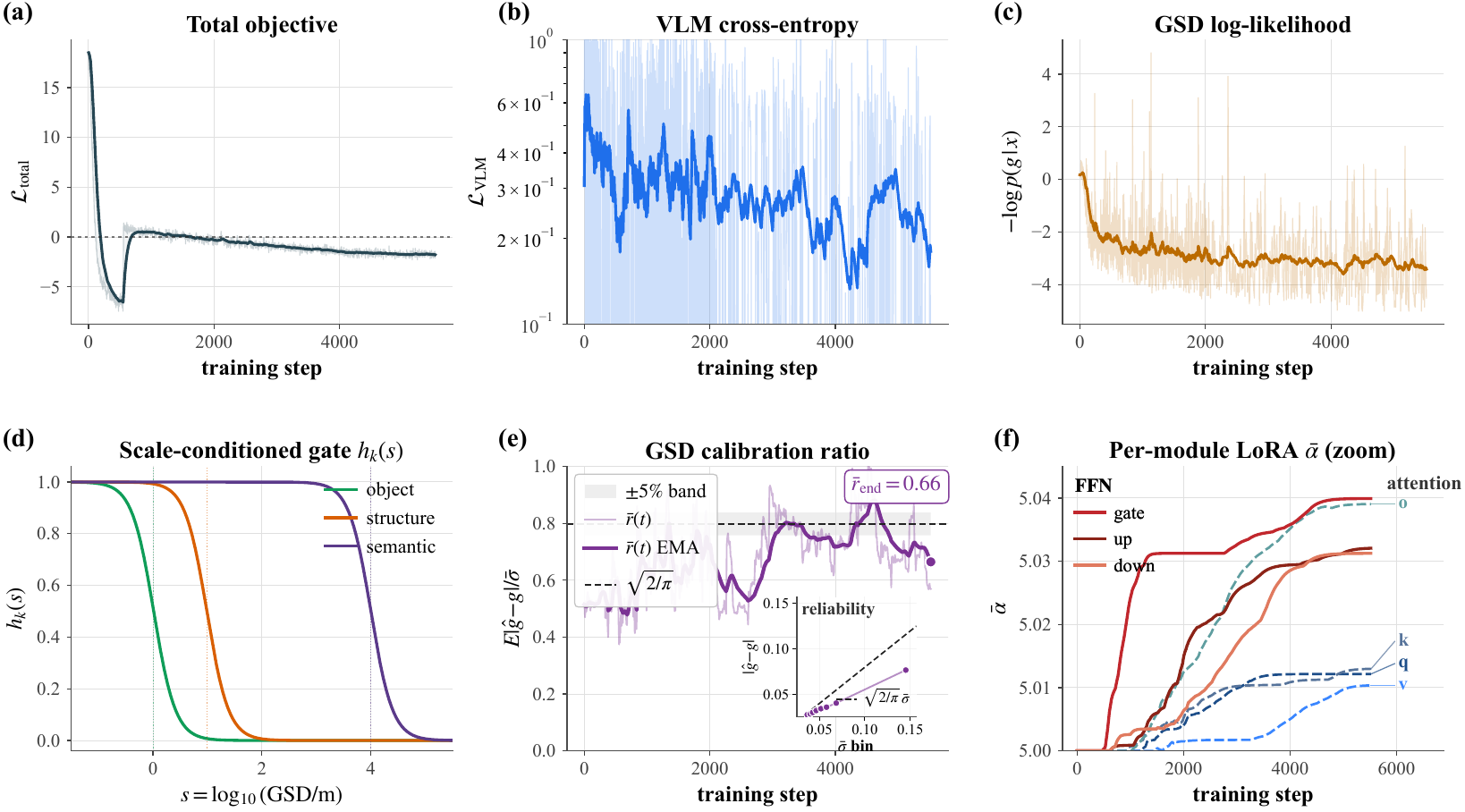}
\caption{\textbf{Stage~2 CS-HLoRA and SSE-U joint-training overview.}
(a)~Total objective
$\mathcal{L}_{\text{total}}
=\mathcal{L}_{\text{VLM}}
+\lambda_{\text{gsd}}^{\text{eff}}\mathcal{L}_{\text{SSE-U}}$.
(b)~VLM cross-entropy loss $\mathcal{L}_{\text{VLM}}$.
(c)~GSD negative log-likelihood $-\log p(g\mid x)$.
(d)~Learned scale-conditioned gates
$h_k(s),\,s{=}\log_{10}(\mathrm{GSD/m})$.
(e)~Calibration ratio $\mathbb{E}|\hat g{-}g|/\bar\sigma$.
(f)~Per-module LoRA effective update magnitude
$\bar\alpha=\frac{\alpha}{r}
\sqrt{\mathrm{tr}(B^\top BAA^\top)}$.}
\label{fig:stage2-overview}
\end{figure*}

\begin{figure*}[t]
\centering
\includegraphics[width=\textwidth]{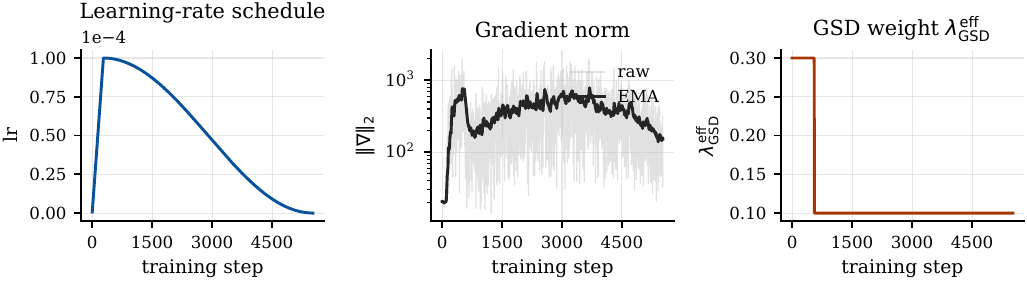}
\caption{\textbf{Stage~2 optimisation knobs.}
\textit{Left:} cosine learning-rate schedule.
\textit{Middle:} gradient norm $\|\nabla\|_2$ on a logarithmic scale.
\textit{Right:} effective GSD-loss weight
$\lambda_{\text{gsd}}^{\text{eff}}$.}
\label{fig:stage2-optim}
\end{figure*}

Figures~\ref{fig:stage1-overview}--\ref{fig:stage2-optim}
summarise the optimisation trajectories of the two training stages.
They provide additional evidence that the training recipe reported in
Tables~\ref{tab:stage1-hp} and~\ref{tab:stage2-hp} is stable and
well-conditioned.

\paragraph{Stage~1 dynamics.}
Figure~\ref{fig:stage1-overview} shows that Stage~1 follows the expected
behaviour of a stable full-parameter SFT run, with rapid initial descent
followed by a low-variance plateau. The SFT loss decreases from
approximately $2.0$ to $0.30$ within the first $800$ steps and then
stabilises around $0.28$--$0.30$, suggesting that one epoch over the
Stage~1 RS corpus is sufficient for domain adaptation. The checkpoint
resume around step~$2000$ does not introduce any visible discontinuity,
confirming that the optimiser and EMA states are restored correctly.
The gradient norm remains bounded throughout training; the isolated
spike at $\|\nabla\|{=}124$ is a benign data-induced outlier and does
not destabilise the loss. Together with the smooth cosine schedule and
linear warmup, these dynamics indicate that the backbone has absorbed
the RS data distribution without obvious optimisation instability. This
is the desired hand-off state for Stage~2.

\paragraph{Stage~2 dynamics and optimisation knobs.}
Figures~\ref{fig:stage2-overview} and~\ref{fig:stage2-optim} jointly
summarise the behaviour of the CS-HLoRA and SSE-U joint-training stage.
The multi-task objective is not a zero-sum trade-off between language
modelling and scale estimation: the VLM cross-entropy decreases
monotonically from approximately $0.6$ to $0.18$, while the GSD
negative log-likelihood also converges throughout training. The early
dip of $\mathcal{L}_{\text{total}}$ below zero is caused by the SSE-U
NLL term once the GSD head becomes confident on precise-GSD batches.
After the effective GSD-loss weight
$\lambda_{\text{gsd}}^{\text{eff}}$ is reduced from $0.3$ to $0.1$
around step~$550$, the SSE-U term contributes less to the total loss
and the optimisation budget shifts toward $\mathcal{L}_{\text{VLM}}$.

The most diagnostic evidence for CS-HLoRA is the learned gate behaviour
in Fig.~\ref{fig:stage2-overview}(d). The three hierarchy gates
self-organise into a clean object, structure, and semantic partition of
scale space: the object branch is active at sub-metre resolutions, the
structure branch around city-block scale, and the semantic branch at
coarser regional scales. This matches the intended inductive bias without
direct supervision on the gate values. The calibration ratio further
shows that SSE-U produces meaningful uncertainty estimates, while the
per-module LoRA effective scale indicates that FFN modules receive
stronger adapter capacity than attention projections during the early
phase of RS instruction tuning. Finally, Fig.~\ref{fig:stage2-optim}
shows that the cosine learning-rate schedule, bounded gradient norm, and
controlled GSD-weight transition keep the joint training in a stable
optimisation regime throughout the run.

\subsection{GSD-Spoofing Test}
\label{app:spoofing}

A common concern is that any LoRA adapter that varies its parameters
with GSD will inherit the benefits of CS-HLoRA simply by adding
\emph{conditional capacity}, regardless of whether the GSD signal is
treated as a continuous physical quantity or merely as a discrete
routing label. To isolate these two factors, we run a controlled
intervention in which the visual input and the question are fixed
while the GSD value provided to the model is systematically perturbed,
and we evaluate four variants that differ only in \emph{how} the GSD
signal is consumed.

\paragraph{Protocol.}
We use the RSVQA-HR test split, whose ground-truth ground sample
distance is $g^{\star}{=}0.15$\,m. At inference time, we replace the
provided GSD with a spoofed pseudo-GSD $\tilde g$ swept over $13$
log-spaced values from $0.01$\,m to $30$\,m, covering severe
underestimation and overestimation of the true scale. For each
$\tilde g$, we keep the image, question, and all other inputs
unchanged, and report VQA accuracy with bootstrap $95\%$ confidence
intervals over a class-balanced subset of $N{=}2{,}000$ QA pairs.
We compare four models trained under the same Stage~2 setting:
\textbf{B2}, a standard LoRA baseline with rank $r{=}64$;
\textbf{B3}, which injects the GSD value as plain text in the
prompt; \textbf{B4} (Bucketed MoE-LoRA), a discrete mixture of
$K{=}3$ LoRA experts hard-routed by the same high/mid/low GSD
buckets that GeoScale-VQA uses, with per-expert rank chosen so
total trainable parameters match CS-HLoRA; and \textbf{CS-HLoRA},
which routes the GSD signal through scale-conditioned hierarchy
gates with analytic sigmoid form. B4 follows the discrete LoRA-MoE
paradigm~\citep{luo2024moelora, dou2023loramoe}, where multiple LoRA
experts are routed by a discrete signal --- in our case the GSD
bucket identity. This baseline lets us separate the contribution
of discrete conditional routing from the continuous physical gating
in CS-HLoRA: B4 inherits the same bucketed structure that
GeoScale-VQA uses for data construction but lacks continuous
interpolation across scales.

\begin{figure*}[t]
\centering
\includegraphics[width=0.7\textwidth]{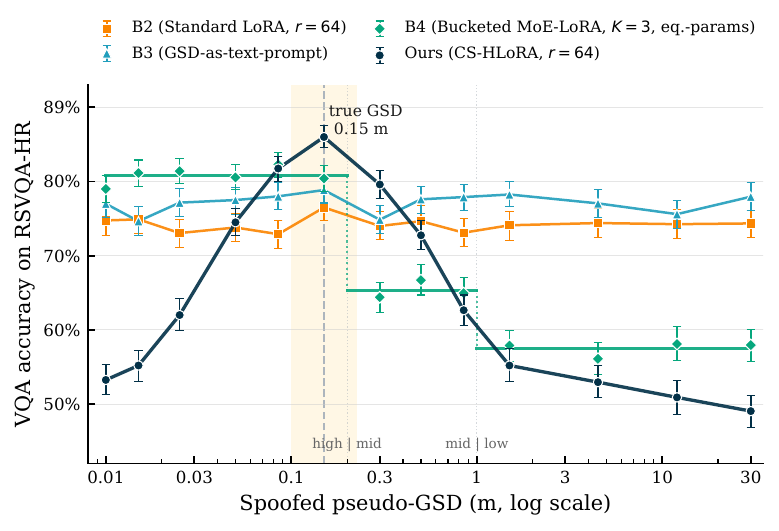}
\caption{\textbf{GSD-spoofing test on RSVQA-HR.}
The true GSD is $0.15$\,m. \textbf{Ours (CS-HLoRA)} peaks near the
true GSD and degrades smoothly under scale mismatch, producing a
clean bell on log-GSD. \textbf{B4 (Bucketed MoE-LoRA)} is
parameter-matched and also peaks in the correct (high) bucket,
confirming that conditional capacity helps; however its response is
piecewise-constant within buckets and discontinuous at bucket edges,
falling well below CS-HLoRA's peak and producing visibly different
predictions for visually near-identical images sitting on opposite
sides of a bucket boundary. \textbf{B2} and \textbf{B3} remain
nearly flat across the entire sweep, indicating that neither standard
LoRA nor textual GSD injection produces genuinely scale-conditioned
inference.}
\label{fig:spoofing}
\end{figure*}

\paragraph{Results.}
Figure~\ref{fig:spoofing} reveals four qualitatively distinct
response patterns. \textbf{CS-HLoRA} achieves its maximum accuracy
near the true GSD, reaching approximately $86\%$ at
$\tilde g{=}0.15$\,m, and decreases smoothly when the provided
scale moves away from the true value, falling to roughly the low
$60\%$ range when $\tilde g$ is close to $1$\,m and approaching
chance-level performance at the largest spoofed values. The left
side of the curve shows a similar degradation under severe scale
underestimation. This bell-shaped response, smooth across the full
two-decade sweep, indicates that the model is not merely using GSD
as a passive token but actively conditions its answer distribution
on the scale signal in a continuous manner.

\textbf{B4} (Bucketed MoE-LoRA) provides the most informative
contrast. Inside the correct \emph{high} bucket where the true GSD
lives, B4 reaches approximately $81\%$, a substantial lift over
B2's $\sim74\%$. This confirms that part of CS-HLoRA's gain can be
attributed to GSD-conditional capacity --- routing different scale
regimes through different parameters helps, even when the routing
mechanism is discrete. However, two distinct deficits remain.
First, B4's in-bucket peak still falls roughly five points short of
CS-HLoRA's $86\%$ peak; even with parameters matched, the discrete
mixture cannot match the per-rank scale-aware modulation that
analytic sigmoid gates provide. Second, B4 produces sharp
discontinuities at bucket edges: spoofing $\tilde g$ from $0.18$\,m
(\emph{high}) to $0.22$\,m (\emph{mid}) drops accuracy by roughly
$15$ points despite the two values being indistinguishable to any
realistic scale estimator. CS-HLoRA, with its smooth gates,
transitions continuously across these same boundaries. This
quantitatively isolates the contribution of the \emph{continuous}
formulation from the contribution of \emph{conditional capacity}:
buckets account for the gap from B2 to B4 ($\sim7$ points
in-bucket), while continuous gating accounts for the additional gap
from B4 to CS-HLoRA ($\sim5$ points in-bucket plus the elimination
of boundary discontinuities).

\textbf{B2} remains approximately flat at $74{-}76\%$ across the
sweep, showing that a standard LoRA adapter of the same rank
largely ignores the GSD channel and answers from the image and
question alone. \textbf{B3} also remains nearly flat, although at a
slightly higher level of $76{-}78\%$, indicating that placing GSD
in the prompt provides a weak textual prior but does not produce
genuine scale-conditioned inference. Together, the contrast between
the smooth bell of CS-HLoRA, the staircase of B4, and the flat
responses of B2 and B3 supports the claim that the performance gain
of CS-HLoRA arises specifically from continuous physical
conditioning, not merely from added parameters, textual injection,
or discrete specialized routing.

This test also provides a practical interpretation of SSE-U
calibration. When the provided or estimated GSD is close to the
true scale, CS-HLoRA routes the sample through the appropriate
hierarchy level and preserves high VQA accuracy. When the scale
estimate deviates substantially, the same image and question can
lead to a markedly different prediction distribution. Accurate and
calibrated scale estimation is therefore not only an auxiliary
objective, but a necessary component for reliable scale-conditioned
decoding.

\subsection{\texorpdfstring{$\tau$-Decoupling Probe}{Tau-Decoupling Probe}}
\label{app:tau}

The spoofing experiment verifies that CS-HLoRA uses the GSD signal at
inference time. We next ask whether the internal rank dimensions induced
by the hierarchy encode distinct visual factors. The gate curves in
Fig.~\ref{fig:stage2-overview}(d) show when different hierarchy levels
are activated along the GSD axis, but they do not by themselves reveal
what those levels represent. We therefore perform a per-rank linear
probing analysis on the CS-HLoRA bottleneck representation.

\paragraph{Protocol.}
We sample $N{=}5{,}000$ probe images covering the GSD range of the
Stage~2 mixture. For each image, we extract the
$64$-dimensional bottleneck latent
$\tau\in\mathbb{R}^{64}$ before the final up-projection. For each rank
dimension $k\in\{0,\dots,63\}$, we fit three independent ridge probes
from the scalar $\tau_k$ to three visual factors:
\begin{enumerate}
\item \textbf{Texture}, measured by Gabor energy.
\item \textbf{Geometry}, measured by Canny edge density.
\item \textbf{Semantics}, measured by source or scene-level identity.
\end{enumerate}
We report the held-out $R^{2}$ of each scalar probe, which measures how
much of each factor is linearly recoverable from an individual rank
dimension.

\begin{figure}[t]
\centering
\includegraphics[width=0.5\linewidth]{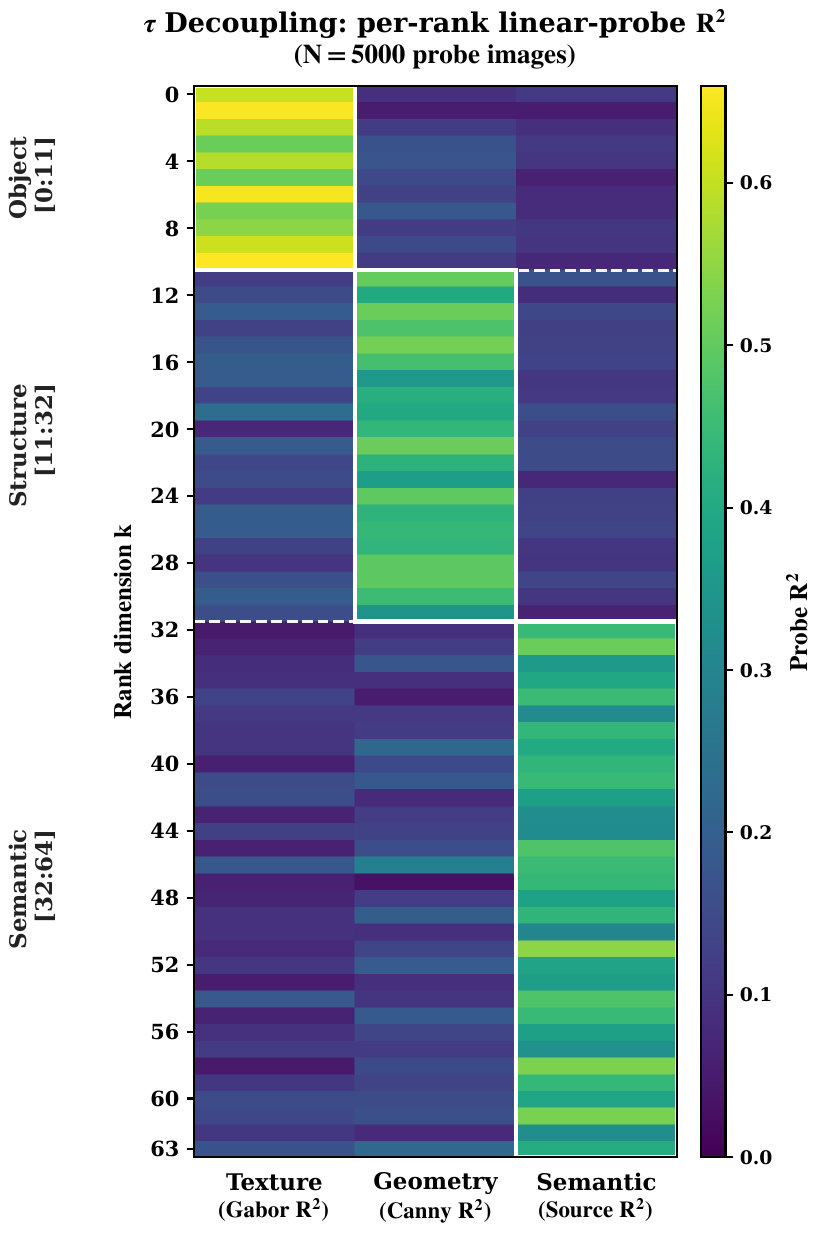}
\caption{\textbf{$\tau$-decoupling analysis.}
Per-rank linear-probe $R^{2}$ on $N{=}5{,}000$ probe images. Rows are
CS-HLoRA rank dimensions grouped into object, structure, and semantic
levels. Columns correspond to texture, geometry, and semantic factors.
The block-diagonal pattern indicates factor-specific specialisation.}
\label{fig:tau}
\end{figure}

\paragraph{Results.}
Figure~\ref{fig:tau} shows a pronounced block-diagonal structure. The
object group, corresponding to rank dimensions $[0{:}11]$, has high
probe $R^{2}$ on the texture column and low values on the geometry and
semantic columns. This is consistent with the role of the object-level
branch, which is most active at fine GSD where local appearance,
high-frequency patterns, and small objects remain visible.

The structure group, corresponding to rank dimensions $[11{:}32]$, has
the strongest response on the geometry probe. This matches the intended
role of the structure-level branch, where buildings, roads, blocks, and
spatial layouts become the dominant cues. The semantic group,
corresponding to rank dimensions $[32{:}64]$, is most predictive of the
semantic or source-level factor, while carrying relatively weak texture
and geometry information. This is consistent with coarse-scale imagery,
where local details are less resolved and scene-level identity becomes
the most reliable signal.

These results complement the GSD-spoofing test. The spoofing curve shows
that CS-HLoRA changes its predictions when the scale signal is perturbed,
while the $\tau$-decoupling heatmap shows that the hierarchy levels encode
different visual factors. Together, the two analyses provide mechanistic
evidence that CS-HLoRA's gains are attributable to scale-aware hierarchical adaptation, rather than to LoRA capacity or prompt-level GSD
conditioning alone.

\section{Qualitative Verification}
\label{appendix:qualitative}

This appendix provides qualitative evidence complementary to the
quantitative results in the main paper. We first examine how different
GSD-injection strategies affect the granularity of model responses, and
then compare ScaleEarth with proprietary VLMs on representative
XLRS-Bench examples.

\subsection{Comparison of GSD Injection Strategies}
\label{appendix:gsd_injection_compare}

We compare three configurations built on the same Stage~1 RS-adapted
backbone: (i) a scale-blind RS-VLM that receives no GSD information,
(ii) a GSD-as-token baseline that prepends \texttt{GSD = X\,m} to the
prompt, and (iii) ScaleEarth, which injects scale through CS-HLoRA and
SSE-U. The comparison focuses on whether each model adapts the
granularity of its description to the physical resolution of the image.

For visualization, we annotate ScaleEarth outputs according to the
CS-HLoRA rank groups they correspond to:
\textcolor{red!50}{object-level details} for ranks $0$ to $10$,
\textcolor{yellow!70!black}{structure-level patterns} for ranks $11$ to
$31$, and \textcolor{green!50!black}{semantic-level abstractions} for
ranks $32$ to $63$. These annotations are used only for qualitative
analysis and are not provided to the model.

\begin{figure}[h]
    \centering
    \includegraphics[width=0.8\linewidth]{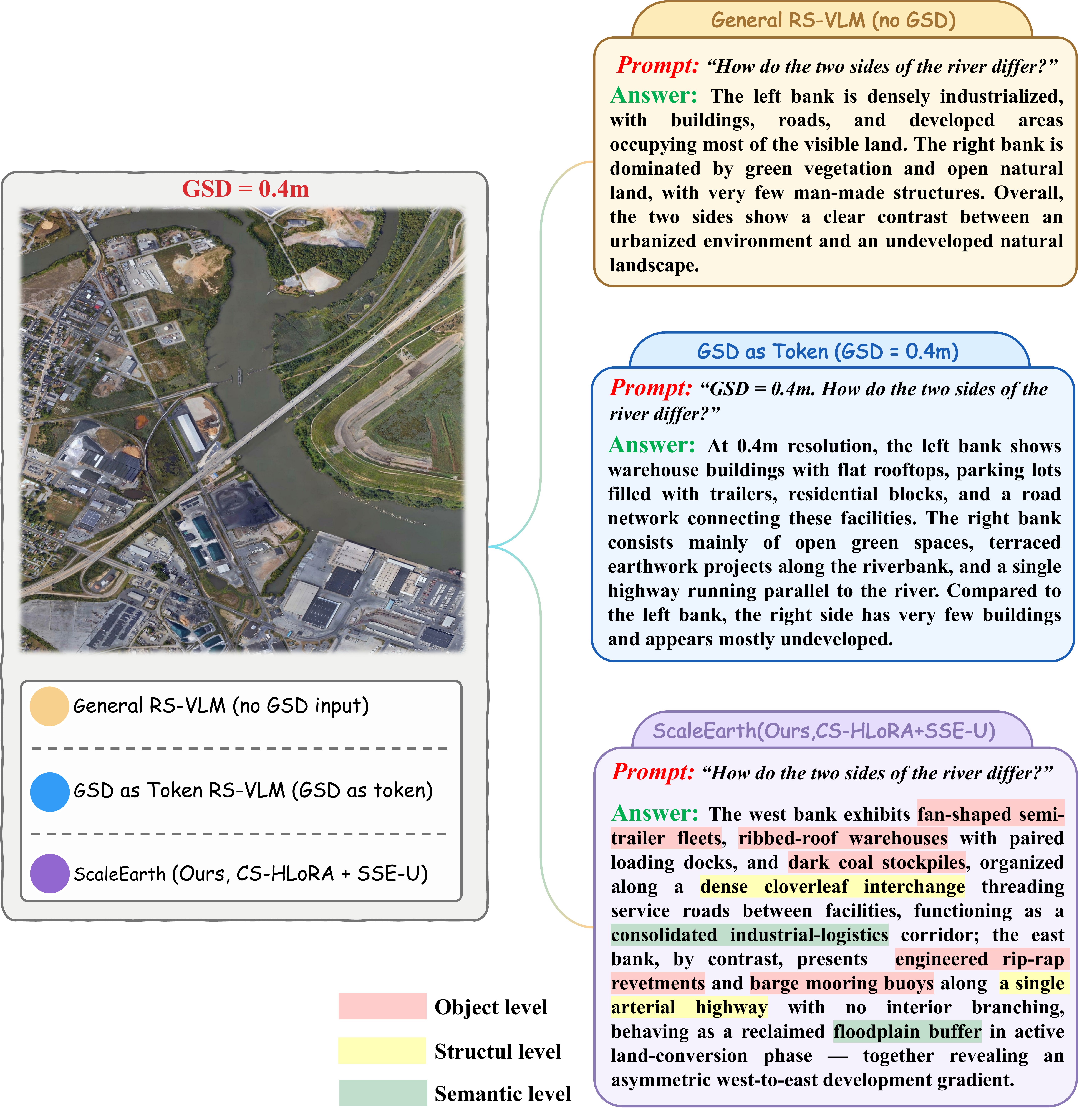}
    \caption{\textbf{Qualitative comparison at GSD $=0.4$\,m.}
    The scale-blind RS-VLM gives a generic urban and natural contrast.
    The GSD-as-token baseline mentions several mid-scale entities.
    ScaleEarth produces a multi-level description containing object-level
    details, structure-level spatial patterns, and semantic-level
    functional interpretation.}
    \label{fig:qual_gsd04}
\end{figure}

\paragraph{High-resolution case.}
Figure~\ref{fig:qual_gsd04} shows a riverbank scene at
GSD $=0.4$\,m. The scale-blind RS-VLM describes the scene using a coarse
urban versus natural contrast, without exploiting fine visual evidence
that is resolvable at this resolution. The GSD-as-token baseline
mentions several visible entities, but its response remains close to a
flat object list. In contrast, ScaleEarth produces a scale-consistent
multi-level description. It identifies object-level cues such as vehicle
fleets, warehouse roofs, stockpiles, revetments, and mooring structures;
structure-level cues such as road connectivity and the spatial separation
between industrial and undeveloped regions; and semantic-level
abstractions such as an industrial logistics corridor and a floodplain
buffer. This behavior is consistent with CS-HLoRA routing, where
fine-scale imagery activates object, structure, and semantic rank groups
jointly.

\begin{figure}[h]
    \centering
    \includegraphics[width=0.8\linewidth]{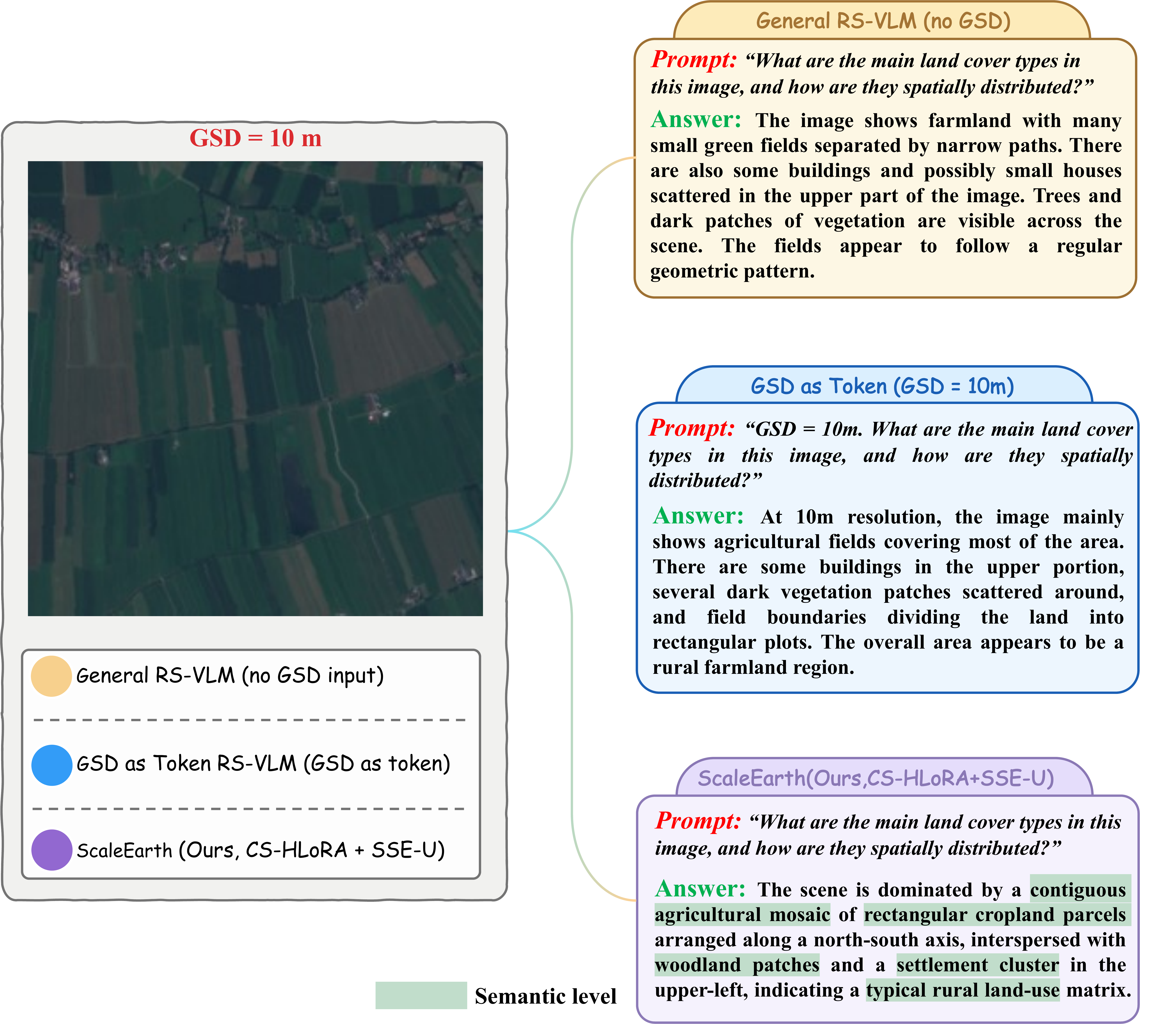}
    \caption{\textbf{Qualitative comparison at GSD $=10$\,m.}
    The scale-blind RS-VLM over-specifies sub-pixel details. The
    GSD-as-token baseline acknowledges the nominal resolution but still
    produces object-level claims. ScaleEarth restricts its response to
    coarse, scale-appropriate semantic descriptions.}
    \label{fig:qual_gsd10}
\end{figure}

\paragraph{Low-resolution case.}
Figure~\ref{fig:qual_gsd10} shows an agricultural scene at
GSD $=10$\,m. At this resolution, individual buildings and narrow paths
are not reliably resolvable. The scale-blind RS-VLM nevertheless
describes such fine-grained entities, indicating that its descriptive
granularity is not calibrated to image resolution. The GSD-as-token
baseline explicitly refers to the nominal resolution, but still produces
object-level claims that are not supported by the image. ScaleEarth
instead focuses on coarse and verifiable concepts, including agricultural
mosaics, rectangular cropland parcels, woodland patches, settlement
clusters, and rural land-use structure. This matches the intended
behavior of CS-HLoRA: when $s=\log_{10}(\mathrm{GSD})$ is large,
object-level ranks are suppressed and the response is dominated by
structure-level and semantic-level evidence.

\paragraph{Discussion.}
These examples show that the three strategies differ not only in factual
correctness, but also in resolution calibration. The scale-blind baseline
tends to describe scenes at a fixed granularity. The GSD-as-token baseline
treats scale primarily as textual context and does not reliably suppress
unsupported fine-grained claims. ScaleEarth modulates its active
representation according to physical resolution, using object-level
vocabulary when the image supports it and suppressing such vocabulary
when the GSD is too coarse.

\subsection{Per-Task Comparison Against Proprietary VLMs}
\label{appendix:per_task_compare}

We further compare ScaleEarth with four proprietary VLMs on five
representative XLRS-Bench examples: Claude Opus~4.7, Gemini~3.1 Pro,
ChatGPT~5.5, and Qwen3.6 Plus. The examples cover object spatial
relationship, object classification, object color, complex reasoning, and
anomaly detection with interpretation. They span three regimes:
localization-dependent perception, coarse semantic perception under
limited effective resolution, and multi-step reasoning that combines
geographic priors with visual evidence.

\begin{figure}[!htbp]
    \centering
    \includegraphics[width=0.7\linewidth]{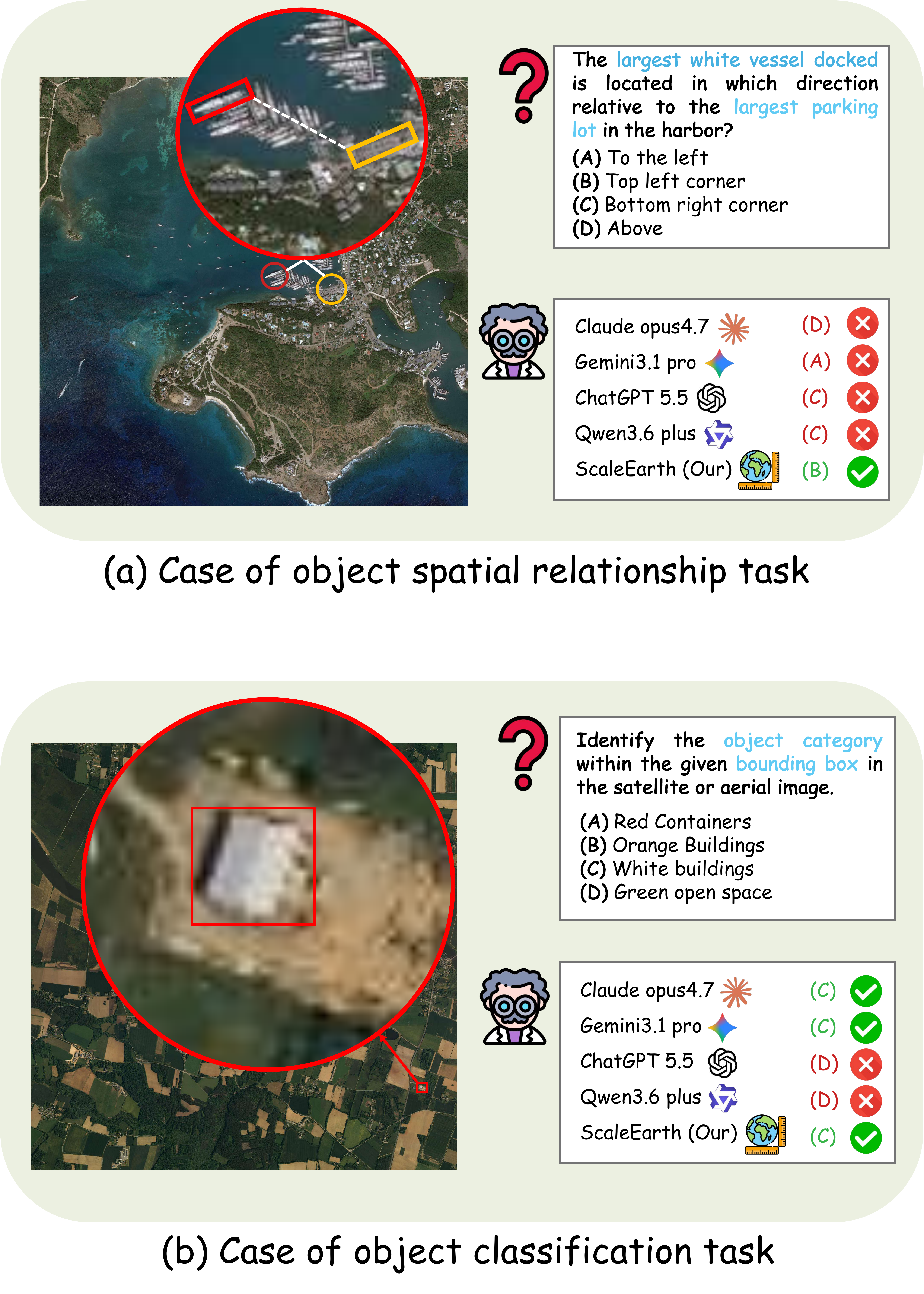}
    \caption{\textbf{Per-task qualitative comparison on XLRS-Bench.}
    (a) Object spatial relationship. ScaleEarth correctly localizes the
    largest white vessel relative to the largest parking lot. The four
    proprietary baselines fail. (b) Object classification in a small
    bounding box. ScaleEarth, Claude Opus~4.7, and Gemini~3.1 Pro
    correctly identify the building category, while ChatGPT~5.5 and
    Qwen3.6 Plus rely on the surrounding agricultural context.}
    \label{fig:qual_taskab}
\end{figure}

\paragraph{Localization-dependent perception.}
Figure~\ref{fig:qual_taskab}(a) requires identifying the largest white
vessel and the largest parking lot in a large harbor scene, followed by
reasoning about their relative position. The proprietary models fail in
different ways, including incorrect reference-frame interpretation and
overly coarse cardinal-direction reasoning. ScaleEarth correctly answers
``top-left corner''. This case illustrates the benefit of scale-aware
grounding: high-resolution training prompts expose the model to dense
small-object localization, and the object-level CS-HLoRA ranks remain
active when the provided GSD supports fine spatial reasoning.

\begin{figure}[!htbp]
    \centering
    \includegraphics[width=0.7\linewidth]{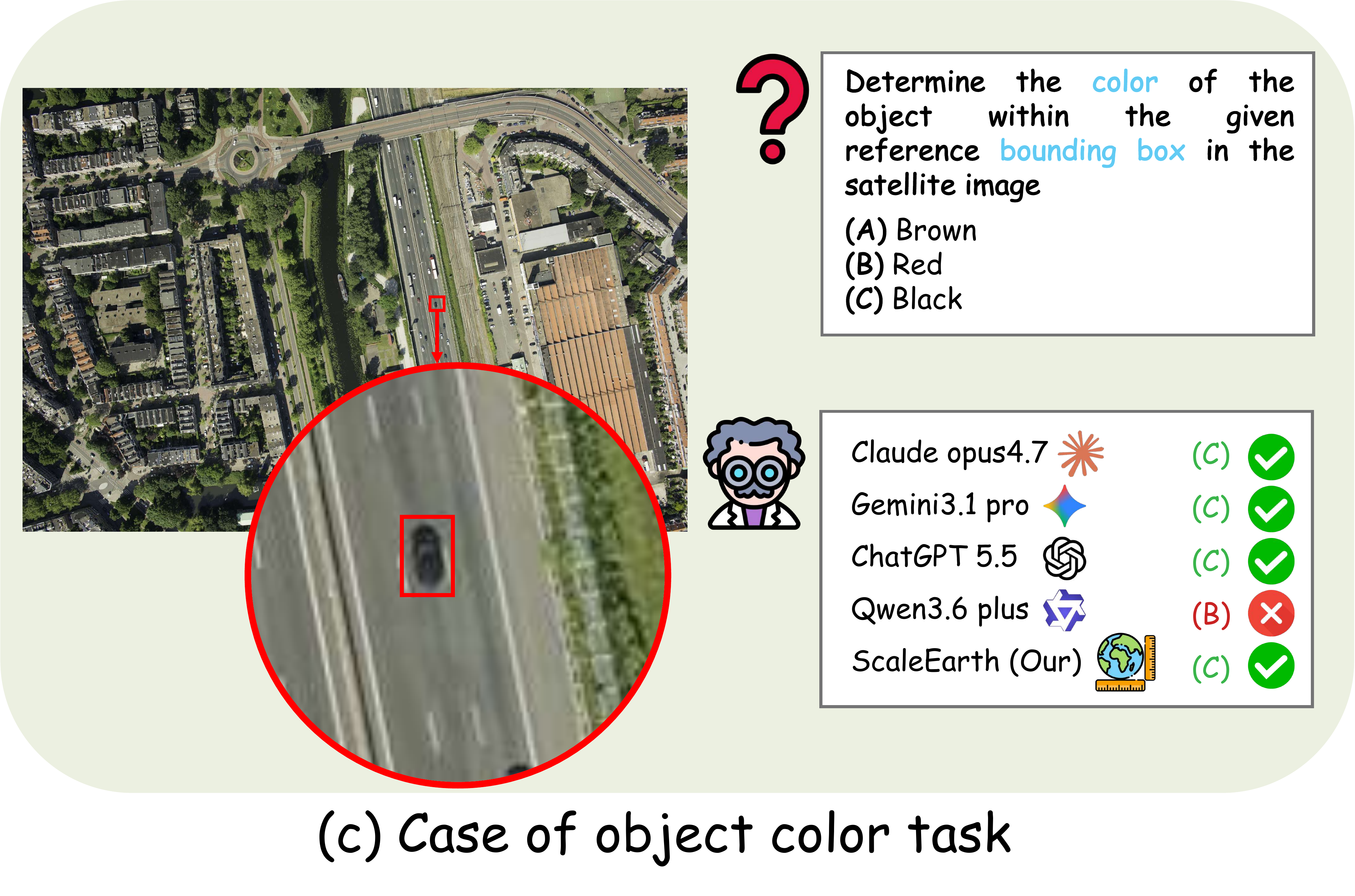}
    \caption{\textbf{Per-task qualitative comparison on XLRS-Bench.}
    (c) Object color in a small, partially shadowed bounding box.
    ScaleEarth, Claude Opus~4.7, Gemini~3.1 Pro, and ChatGPT~5.5
    correctly identify the target vehicle as black, while Qwen3.6 Plus
    misclassifies the hue under shadow.}
    \label{fig:qual_taskc}
\end{figure}

\paragraph{Coarse semantic perception.}
Figure~\ref{fig:qual_taskab}(b) shows a small white-roofed building
surrounded by agricultural land. ChatGPT~5.5 and Qwen3.6 Plus select the
surrounding land-cover category rather than the target inside the
bounding box, while ScaleEarth, Claude Opus~4.7, and Gemini~3.1 Pro
correctly answer ``White buildings''. Figure~\ref{fig:qual_taskc} gives a
more fine-grained color-recognition example, where the target vehicle is
small and partially shadowed. ScaleEarth correctly answers ``Black'', as
do three of the four proprietary models. These cases suggest that
ScaleEarth remains competitive with frontier proprietary VLMs on
localized perception, even when the task mainly depends on structure-level
rather than purely object-level evidence.

\begin{figure}[!htbp]
    \centering
    \includegraphics[width=0.7\linewidth]{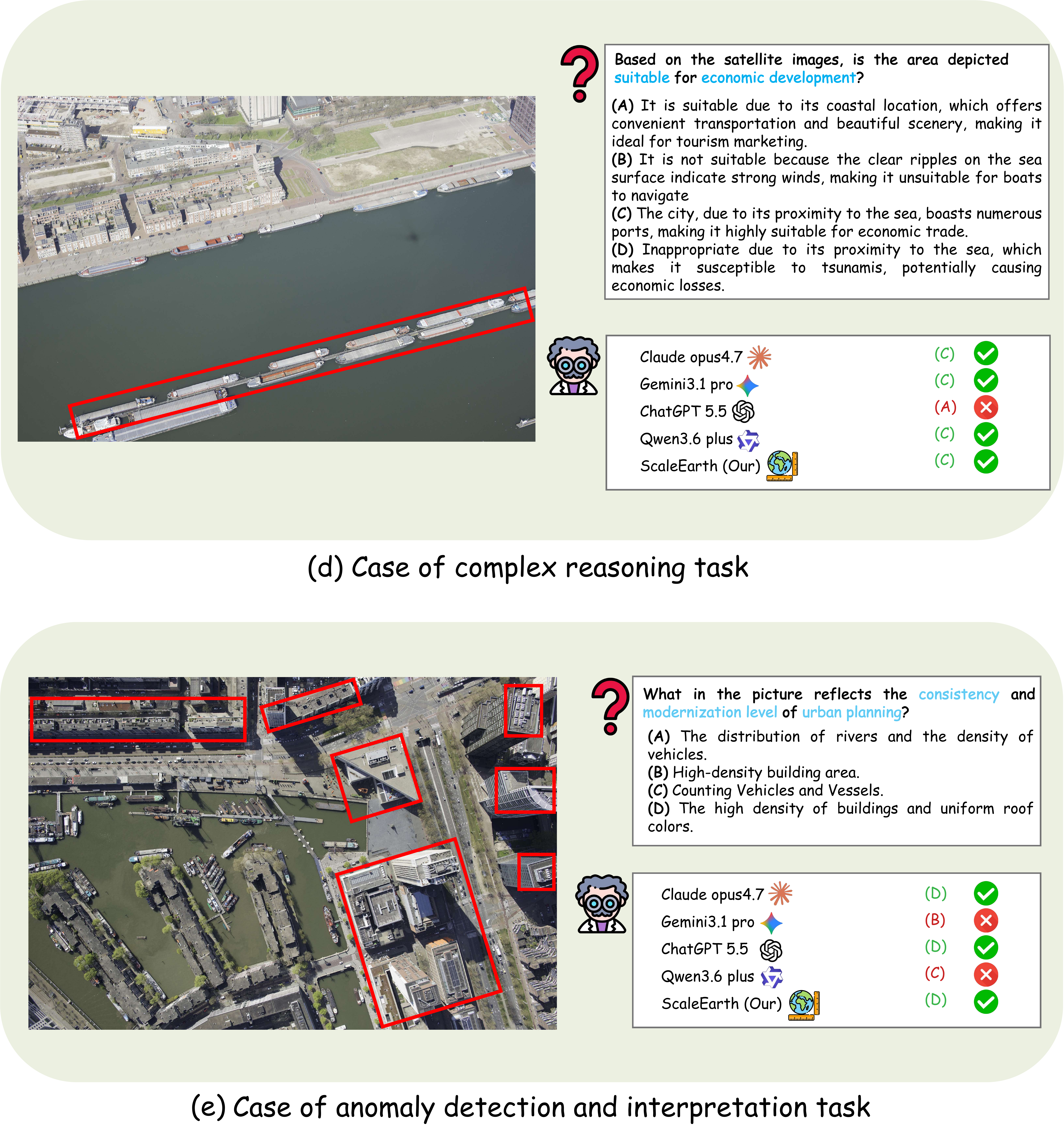}
    \caption{\textbf{Per-task qualitative comparison on XLRS-Bench.}
    (d) Complex reasoning. ScaleEarth identifies the visible barge fleet
    and port infrastructure as evidence of economic-trade suitability.
    (e) Anomaly detection and interpretation. ScaleEarth identifies the
    joint cue of roof-color uniformity and building density.}
    \label{fig:qual_taskde}
\end{figure}

\paragraph{Multi-step reasoning.}
Figure~\ref{fig:qual_taskde}(d) asks whether a river-port scene is
suitable for economic development. The correct answer requires connecting
visible barges and port infrastructure to established trade activity,
rather than relying on distractors related to tourism, weather, or natural
hazards. ScaleEarth answers correctly, together with Claude Opus~4.7,
Gemini~3.1 Pro, and Qwen3.6 Plus, while ChatGPT~5.5 selects the tourism
distractor. Figure~\ref{fig:qual_taskde}(e) asks which visual cue best
reflects urban-planning consistency. The correct answer requires jointly
considering building density and roof-color uniformity. ScaleEarth,
Claude Opus~4.7, and ChatGPT~5.5 identify the joint cue correctly, while
Gemini~3.1 Pro and Qwen3.6 Plus miss part of the required evidence. These
examples indicate that the semantic-level rank group supports reasoning
over scene-level structure without sacrificing general visual reasoning
ability.

\paragraph{Discussion.}
Across the five representative tasks, ScaleEarth returns the correct
answer in all cases, while each proprietary baseline fails on at least
one example. The failures of proprietary models are concentrated in
scale-sensitive grounding and localization, which are precisely the
settings targeted by CS-HLoRA. On tasks that rely more heavily on
semantic reasoning than fine-scale grounding, ScaleEarth remains
competitive with substantially larger proprietary models. This suggests
that scale-conditioned adaptation improves resolution-aware grounding
without degrading broader reasoning behavior.

\section{Training Recipe and Dataset Composition}
\label{app:training-recipe}

This appendix provides the complete training recipe for the two-stage
CS-HLoRA pipeline used in the main experiments. Stage~1 performs
full-parameter remote-sensing (RS) domain adaptation, while Stage~2
freezes the RS-adapted backbone and jointly trains the CS-HLoRA adapters
and the SSE-U scale-estimation head. All experiments start from
Qwen3-VL-8B-Instruct and use DeepSpeed ZeRO-3 with bf16 precision.
Both stages are trained on $4$ nodes with $4$ NVIDIA A100 80\,GB GPUs
per node, resulting in $16$ GPUs in total. The wall-clock training time
is approximately \textbf{24\,h} for Stage~1 and \textbf{40\,h} for Stage~2.

\subsection{Stage~1: Full-parameter RS Domain SFT}
\label{app:stage1}

\paragraph{Goal.}
The goal of Stage~1 is to adapt the Qwen3-VL backbone to the visual
statistics, object scales, and instruction style of overhead imagery
before introducing any scale-aware modules. This stage does not use GSD
signals, LoRA adapters, CS-HLoRA, or SSE-U. It serves as a full-parameter
RS supervised fine-tuning stage on which the subsequent scale-conditioned
training is built.

\paragraph{Data.}
Stage~1 uses RS-GPT4V-style remote-sensing instruction data, consisting
of long-form captions, detailed image descriptions, and complex reasoning
instructions over heterogeneous public RS imagery. Two splits are
concatenated and shuffled at each epoch.

\begin{table}[h]
\centering
\small
\renewcommand{\arraystretch}{1.15}
\setlength{\tabcolsep}{7pt}
\begin{tabularx}{0.88\linewidth}{>{\raggedright\arraybackslash}X r >{\raggedright\arraybackslash}X}
\toprule
\textbf{Split} & \textbf{\# Samples} & \textbf{Data Type} \\
\midrule
RS-GPT4V
& $\sim$938\,K
& captions, detailed descriptions, reasoning instructions \\
RS-GPT4V-Instruct
& $\sim$19\,K
& multi-turn instruction-following samples \\
\midrule
\textbf{Total}
& $\sim$957\,K
& mixed RS captioning and instruction data \\
\bottomrule
\end{tabularx}
\caption{Stage~1 training data composition.}
\label{tab:stage1-data}
\end{table}

\paragraph{Trainable parameters.}
All major components are updated in Stage~1, including the vision tower,
the multimodal projector, and the language model. We assign a smaller
learning rate to the vision tower to preserve its pre-trained low-level
visual features while still allowing adaptation to RS imagery.

\begin{table}[h]
\centering
\small
\renewcommand{\arraystretch}{1.15}
\setlength{\tabcolsep}{7pt}
\begin{tabularx}{0.92\linewidth}{>{\raggedright\arraybackslash}p{0.38\linewidth} >{\raggedright\arraybackslash}X}
\toprule
\textbf{Item} & \textbf{Setting} \\
\midrule
Backbone & Qwen3-VL-8B-Instruct \\
Tunable modules & vision tower, multimodal projector, and LLM \\
Scale-aware modules & LoRA, CS-HLoRA, and SSE-U disabled \\
Optimizer & AdamW with $\beta_1{=}0.9$, $\beta_2{=}0.999$ \\
Learning rate for LLM and projector & $1{\times}10^{-5}$ \\
Learning rate for vision tower & $1{\times}10^{-6}$ \\
Learning-rate schedule & cosine decay with warmup ratio $0.03$ \\
Weight decay & $1{\times}10^{-3}$ \\
Gradient clipping & $1.0$ \\
Training epochs & $1$ \\
Per-device batch size & $1$ \\
Gradient accumulation & $8$ \\
Effective global batch size & $1{\times}8{\times}16 = 128$ \\
Sequence length & $8192$ \\
Image pixel range & $[3{,}136,\;12{,}845{,}056]$ \\
Precision & bf16 with gradient checkpointing \\
Distributed training & DeepSpeed ZeRO-3 \\
Hardware & $4$ nodes $\times$ $4$ A100 80\,GB GPUs \\
Wall-clock time & $\approx 24$\,h \\
\bottomrule
\end{tabularx}
\caption{Stage~1 full-parameter RS domain adaptation recipe.}
\label{tab:stage1-hp}
\end{table}

The maximum image budget of $12{,}845{,}056$ pixels
($\approx 3584{\times}3584$) is selected to preserve high-resolution
aerial scenes without overly aggressive downsampling. This is important
for fine-grained RS targets such as small vehicles, ships, roads, and
compact man-made structures. The $8192$-token context length is sufficient
for one high-resolution RS image, up to approximately $6.5$K vision tokens,
together with long-form textual supervision.

\subsection{Stage~2: CS-HLoRA + SSE-U Joint Training}
\label{app:stage2}

\paragraph{Goal.}
Stage~2 injects scale awareness into the Stage~1 RS-adapted backbone.
The backbone is frozen, and only the proposed CS-HLoRA adapters and the
SSE-U scale-estimation head are optimized. The adapters provide
continuous scale-conditioned modulation of the language model, while
SSE-U estimates the effective image scale when reliable metadata is not
available. The training objective is
\[
\mathcal{L}
=
\mathcal{L}_{\text{VLM}}
+
\lambda_{\text{gsd}}
\mathcal{L}_{\text{SSE-U}}^{\text{NLL}},
\]
where $\lambda_{\text{gsd}}{=}0.1$ in the steady state. During the first
$10\%$ of training steps, this weight is increased to $0.3$ to warm up
the SSE-U head before its predictions are heavily coupled with
scale-conditioned language modeling.

\paragraph{Data.}
Stage~2 mixes four RS instruction sources with complementary scale and
reasoning properties. RSVQA provides the dominant VQA supervision,
MtSCCD-VQA contributes multi-temporal and multi-resolution samples,
PatternNet-VQA strengthens fine-grained scene recognition, and
GeoLLaVA-8K provides geographic and reasoning-oriented instruction data.
GeoLLaVA-8K is included only in Stage~2, so that geographic reasoning is
introduced after the backbone has acquired basic RS visual-language
literacy and when scale-conditioned training becomes active.

\begin{table}[h]
\centering
\small
\renewcommand{\arraystretch}{1.15}
\setlength{\tabcolsep}{5.2pt}
\begin{tabularx}{0.98\linewidth}{
>{\raggedright\arraybackslash}p{0.27\linewidth}
r
>{\raggedright\arraybackslash}p{0.21\linewidth}
>{\raggedright\arraybackslash}X}
\toprule
\textbf{Source} & \textbf{\# Samples} & \textbf{GSD Annotation} & \textbf{Role} \\
\midrule
RSVQA family
& $\sim$1.14\,M
& precise
& dominant VQA supervision from LR, HR, and xBEN subsets \\
MtSCCD-VQA
& $\sim$194\,K
& precise
& multi-temporal urban-scene reasoning at known acquisition scale \\
PatternNet-VQA
& $\sim$77\,K
& precise
& fine-grained scene-category and object-level recognition \\
GeoLLaVA-8K
& $\sim$80\,K
& range-based
& geographic understanding and reasoning-oriented instruction tuning \\
\midrule
\textbf{Total}
& $\sim$1.49\,M
& mixed
& scale-aware RS VQA and reasoning supervision \\
\bottomrule
\end{tabularx}
\caption{Stage~2 training data composition. RSVQA family includes the
LR, HR, and xBEN subsets. Most sources provide precise GSD annotations,
while GeoLLaVA-8K provides a coarse GSD range and is handled through
range-based scale inference.}
\label{tab:stage2-data}
\end{table}

To prevent the scale gate from overfitting to sources with exact GSD
metadata, we use a \textbf{BalancedScaleSampler}. Each mini-batch is
constrained to contain an exact-GSD ratio of $0.25$ and at least two
samples from distinct scale bins. For range-annotated samples, we draw a
scale value from the corresponding GSD interval during training. In
addition, precise GSD values are replaced by the SSE-U estimate with probability $p_{\text{e2e}}{=}0.2$, which creates a curriculum bridge
between exact metadata and model-estimated scale.

\paragraph{Trainable parameters.}
Only the CS-HLoRA adapters and the SSE-U head are trainable in Stage~2.
The vision tower, multimodal projector, and LLM backbone remain frozen at
their Stage~1 values. CS-HLoRA uses rank $r{=}64$, scaling factor
$\alpha_{\mathrm{lora}}{=}32$, and zero dropout, and is attached to all
attention projections (q/k/v/o) and FFN projections (gate/up/down) of the
LLM blocks; it is not inserted into the ViT blocks.

\begin{table}[h]
\centering
\small
\renewcommand{\arraystretch}{1.15}
\setlength{\tabcolsep}{7pt}
\begin{tabularx}{0.94\linewidth}{>{\raggedright\arraybackslash}p{0.42\linewidth} >{\raggedright\arraybackslash}X}
\toprule
\textbf{Item} & \textbf{Setting} \\
\midrule
Initialization & Stage~1 checkpoint \\
Trainable modules & CS-HLoRA adapters and SSE-U head \\
CS-HLoRA placement & LLM blocks only: q, k, v, o, gate, up, down; no CS-HLoRA in the ViT \\
Frozen modules & vision tower, multimodal projector, and LLM backbone \\
Backbone storage precision & bf16 (primary); 4-bit NF4 for the efficiency variant \\
CS-HLoRA rank / scaling factor $\alpha_{\mathrm{lora}}$ & $64$ / $32$ \\
CS-HLoRA dropout & $0$ \\
Adapter learning rate & $1{\times}10^{-4}$ \\
SSE-U learning rate & $1{\times}10^{-4}$ \\
SSE-U replacement probability & $p_{\text{e2e}}{=}0.2$ \\
SSE-U loss weight & $\lambda_{\text{gsd}}{=}0.1$, increased to $0.3$ during the first $10\%$ steps \\
BalancedScaleSampler exact-GSD ratio & $0.25$ \\
BalancedScaleSampler minimum scale bins & $2$ \\
Optimizer & AdamW \\
Learning-rate schedule & cosine decay with warmup ratio $0.05$ \\
Weight decay & $1{\times}10^{-3}$ \\
Gradient clipping & $5.0$ \\
Training epochs & $1$ \\
Per-device batch size & $1$ \\
Gradient accumulation & $16$ \\
Effective global batch size & $1{\times}16{\times}16 = 256$ \\
Sequence length & $16{,}384$ \\
Image pixel range & $[3{,}136,\;12{,}845{,}056]$ \\
Precision & bf16 with gradient checkpointing \\
Distributed training & DeepSpeed ZeRO-3 with NCCL timeout $7200$\,s \\
Hardware & $4$ nodes $\times$ $4$ A100 80\,GB GPUs \\
Wall-clock time & $\approx 40$\,h \\
\bottomrule
\end{tabularx}
\caption{Stage~2 CS-HLoRA and SSE-U joint-training recipe.}
\label{tab:stage2-hp}
\end{table}

The sequence length is increased to $16{,}384$ in Stage~2 to accommodate
both high-resolution visual inputs and the additional scale-related
tokens consumed by the SSE-U pathway. In practice, this allows the model
to process images near the maximum pixel budget while preserving enough
context for multi-turn RS questions and long-form reasoning. The larger
gradient clipping threshold and higher warmup ratio are used because
Stage~2 updates only a small fraction of the total parameters; overly
tight clipping would suppress adapter learning and slow down convergence.

\section{Evaluation Benchmark Details}
\label{app:benchmarks}

This appendix provides additional details on the two remote-sensing
benchmarks used in our evaluation. We summarize their data sources,
evaluation taxonomies, annotation procedures, and scoring protocols, and
clarify why the two benchmarks provide complementary evidence for
assessing ScaleEarth.

\subsection{XLRS-Bench}
\label{app:xlrs}

\textbf{Motivation and Scope.}
XLRS-Bench~\citep{wang2025xlrs} is designed to evaluate whether
multimodal large language models can understand native
ultra-high-resolution (UHR) remote-sensing scenes. This setting differs
substantially from conventional RS-VQA benchmarks, which typically rely
on cropped image tiles and therefore only expose models to local spatial
contexts. By contrast, XLRS-Bench evaluates models on full UHR images,
with an average spatial size of approximately
$8{,}500{\times}8{,}500$ pixels and 840 images at
$10{,}000{\times}10{,}000$ pixels. The benchmark therefore stresses two
capabilities that are central to real-world RS interpretation:
fine-grained perception of small objects and long-range reasoning over
large spatial extents.

\textbf{Image Sources.}
XLRS-Bench contains 1{,}400 real-world UHR remote-sensing images. The
images are collected from existing detection and segmentation datasets
and are selected to cover diverse land-use categories, object densities,
and scene layouts. Specifically, the detection-oriented subset includes
images from DOTA-v2 and ITCVD, while the segmentation-oriented subset
draws from MiniFrance, Toronto, Potsdam, and HRSCD. The resulting image
pool includes both single-temporal UHR scenes and bi-temporal image
pairs, enabling evaluation of static perception, spatial reasoning, and
change-related reasoning within a unified benchmark.

\textbf{Evaluation Taxonomy.}
The benchmark adopts a hierarchical evaluation taxonomy. At the highest
level, XLRS-Bench distinguishes between \emph{perception} and
\emph{reasoning}. These are further decomposed into capability families
and finally into 16 concrete sub-tasks. The VQA-style sub-tasks cover:
\begin{itemize}[leftmargin=*]
    \item \textbf{Counting}: Overall Counting (OC) and Regional Counting (RC).
    \item \textbf{Scene Classification}: Overall Land-Use Classification
    (OLUC) and Regional Land-Use Classification (RLUC).
    \item \textbf{Object Spatial Relationship}: Object Spatial Relationship
    (OSR).
    \item \textbf{Object Properties}: Object Classification (OCC),
    Object Color (OCL), and Object Motion State (OMS).
    \item \textbf{Anomaly Reasoning}: Anomaly Detection and Interpretation
    (AD).
    \item \textbf{Complex Reasoning}: Environmental Condition Reasoning
    (ECR) and Counting with Complex Reasoning (CCR).
    \item \textbf{Planning}: Route Planning (RP).
    \item \textbf{Spatiotemporal Reasoning}: Regional Counting with
    Change Detection (RCCD).
\end{itemize}
In addition to the above VQA-style dimensions, XLRS-Bench includes
captioning and visual-grounding tasks, which are evaluated with separate
task-specific protocols.

\textbf{Annotation and Quality Control.}
XLRS-Bench emphasizes human-verified annotation. VQA and grounding
samples are manually annotated and cross-validated by multiple annotator
groups, with disagreements further reviewed by an expert team. For
detailed image captioning, the benchmark uses a semi-automated
captioning pipeline tailored to UHR imagery: each large image is
decomposed into local views together with a downsampled global view,
after which candidate captions are generated and manually verified. The
released benchmark contains 45{,}942 annotations, including VQA pairs,
visual-grounding instances, and detailed captions. The benchmark supports
both English and Chinese evaluation.

\textbf{Evaluation Protocol.}
For VQA tasks, questions are formulated as multiple-choice questions and
scored by accuracy. Object Motion State is evaluated as a binary
yes--no task, while the remaining VQA sub-tasks use multi-option
classification. In our evaluation, we report aggregated results at the
capability-family level as the main benchmark scores and provide
per-sub-task results where appropriate. This reporting choice reduces
variance from individual sub-tasks while preserving diagnostic
information about specific perception and reasoning capabilities.

\subsection{OmniEarth-Bench}
\label{app:omniearth}

\textbf{Motivation and Scope.}
OmniEarth-Bench~\citep{wang2025omniearth} evaluates multimodal models
from the perspective of Earth-system cognition rather than remote
sensing scene understanding alone. It covers all six Earth-science
spheres---atmosphere, lithosphere, oceansphere, cryosphere, biosphere,
and human-activity sphere---as well as cross-sphere interactions. This
design makes OmniEarth-Bench complementary to XLRS-Bench. Whereas
XLRS-Bench stresses UHR spatial perception and reasoning over large
remote-sensing scenes, OmniEarth-Bench tests whether a model can handle
heterogeneous observational modalities, domain-specific scientific
concepts, and interactions across Earth-system components.

\textbf{Data Sources.}
OmniEarth-Bench integrates observational data from 33 native sources,
including satellite imagery, multispectral products, weather and climate
records, sea-ice products, seismic data, and in-situ measurements.
Representative sources span multiple Earth-science domains: for example,
human-activity and land-cover data, vegetation and ecosystem monitoring
products, atmospheric and severe-weather datasets, lithospheric and
seismic observations, oceanographic measurements, cryospheric products,
and cross-sphere datasets for compound environmental phenomena. This
heterogeneous design is intended to preserve the diversity of real
Earth-observation data rather than reducing all inputs to a single
remote-sensing image format.

\textbf{Evaluation Taxonomy.}
OmniEarth-Bench organizes its annotations into a four-level hierarchy:
\emph{Sphere}, \emph{Scenario}, \emph{Ability}, and \emph{Task}. The
sphere level specifies the Earth-system component or cross-sphere
setting. The scenario level defines expert-approved application contexts
within each sphere, such as disaster monitoring, ecological assessment,
urban analysis, marine events, or cryospheric observation. The ability
level evaluates four types of model competence: Perception, General
Reasoning, Scientific-Knowledge Reasoning, and Chain-of-Thought (CoT)
Reasoning. The task level instantiates these dimensions into 109
fine-grained expert-curated tasks.

\textbf{Annotation and Quality Control.}
The benchmark follows a top-down construction strategy. Domain experts
first define evaluation dimensions grounded in real Earth-science
problems, after which data are curated or collected to populate the
corresponding tasks. The annotation process combines expert design,
manual question construction, and human verification. In particular,
OmniEarth-Bench involves experts from relevant Earth-science domains
together with crowd-sourced annotators, and applies hybrid
expert--crowd validation to reduce label ambiguity. For CoT-style
samples, reference reasoning chains are checked and revised to ensure
that the final explanations remain logically grounded rather than merely
template-generated.

\textbf{Question Formats and Protocols.}
OmniEarth-Bench supports multiple evaluation formats:
\begin{itemize}[leftmargin=*]
    \item \textbf{Multiple-Choice VQA}: questions are evaluated by
    classification accuracy. The option set includes a correct answer,
    distractors, and an uncertainty option to discourage unsupported
    guessing.
    \item \textbf{Open-Ended VQA}: prompts are converted into free-form
    generation and judged against expert-written reference answers.
    \item \textbf{Visual Grounding}: localization outputs are evaluated
    using precision at predefined IoU thresholds.
    \item \textbf{Chain-of-Thought Reasoning}: reasoning traces are
    assessed for both informativeness and factual correctness, following
    established MLLM reasoning-evaluation protocols.
    \item \textbf{Image Captioning}: generated descriptions are evaluated
    with standard captioning metrics, including BLEU, METEOR,
    ROUGE-L, and CIDEr.
\end{itemize}

\textbf{Complementarity in Our Evaluation.}
The two benchmarks probe different but complementary aspects of
ScaleEarth. XLRS-Bench provides a controlled stress test for native UHR
remote-sensing interpretation, where scale awareness is expected to be
most directly beneficial because object visibility, spatial context, and
semantic granularity all depend strongly on GSD. OmniEarth-Bench, in
contrast, evaluates whether the model maintains broad Earth-system
competence across heterogeneous data modalities and cross-sphere
scientific reasoning. Strong performance on both benchmarks therefore
indicates not only improved scale-conditioned RS perception, but also
preserved generality across diverse Earth-observation tasks.

\section{Related Work}
\label{sec:related_work}

\paragraph{Remote Sensing Vision--Language Models.}
The extension of VLMs to Earth observation has progressed rapidly. GeoChat~\citep{kuckreja2024geochat} builds on LLaVA-style
architectures with large-scale region-based instruction data,
establishing a representative template for RS-VLM design;
EarthGPT~\citep{zhang2024earthgpt} scales this recipe to
multi-sensor inputs with more than one million samples; and a series of
follow-ups~\citep{luo2024skysensegpt,zhan2025skyeyegpt}
further extend fine-grained relational reasoning and instruction diversity. Recent systems push two frontiers: pixel-level grounding, and ultra-high-resolution (UHR) perception, where
GeoLLaVA-8K~\citep{wang2025geollava} targets 8K-scale inputs.
Unlike prior RS-VLMs that treat scale as an implicit or discrete attribute, \textbf{ScaleEarth} explicitly models GSD as a \emph{continuous} physical variable that modulates parameter
activation, bridging the gap between spatial resolution and semantic
representation.

\vspace{-0.8em}
\paragraph{Parameter-Efficient Fine-Tuning for RS-VLMs.}
Adapting billion-scale VLMs to RS imagery is constrained by memory and computation. LoRA~\citep{hu2022lora} has therefore become a widely adopted recipe across GeoChat, SkySenseGPT, and most follow-ups, while
QLoRA~\citep{dettmers2023qlora} further reduces the adaptation footprint by backpropagating through a frozen 4-bit quantized backbone into
low-rank adapters. Recent work also studies quantization-aware training for efficient VLMs: GRACE~\citep{chen2026grace} unifies knowledge
distillation and QAT under an information-bottleneck perspective, showing that low-bit VLMs can retain strong multimodal performance while
substantially reducing deployment cost. These results suggest that low-bit adaptation is a promising direction for deployable VLMs, but
existing quantization-oriented methods are not designed around the physical scale variation that dominates overhead imagery. A parallel line compresses the visual-token stream rather than the weights:
LLaVA-UHD~\citep{guo2024llava} and Monkey~\citep{li2024monkey}
compress patch tokens in the general domain, while
GeoLLaVA-8K~\citep{wang2025geollava} observes that UHR imagery is dominated by semantically sparse background and proposes Background Token Pruning with Anchored Token Selection. Conditional-LoRA variants such as MoELoRA~\citep{luo2024moelora} and MoCLE~\citep{gou2026mixture} add a discrete mixture of expert routing, but the routing signal is inferred from input content and does not offer an interpretable physical handle. Two gaps emerge: quantization and token pruning improve efficiency but apply identical adapter parameters across resolutions, whereas discrete-expert routing exposes no physical variable to the user. \textbf{CS-HLoRA} modulates the low-rank subspace through a smooth, differentiable gate driven by GSD, and its 4-bit configuration preserves the QLoRA memory profile. QLoRA trades a small amount of precision for efficiency relative to LoRA; we report it because ultra-high-resolution inputs make memory the binding constraint, and every baseline is adapted in bf16, so the 4-bit CS-HLoRA competes while absorbing the quantization penalty rather than benefiting from it. The bf16 row of Table~\ref{tab:xlrs_main} shows the method without that penalty.

\vspace{-0.8em}
\paragraph{RS-VQA Datasets and GSD in Instruction Data.}
RSVQA~\citep{lobry2020rsvqa} automatically generates question--answer
pairs from RS datasets at two predefined resolutions (LR and HR);
GeoChat-Instruct~\citep{kuckreja2024geochat} adds region-grounded
samples; FIT-RS~\citep{luo2024skysensegpt} contributes scene-graph
relational VQA; and SuperRS-VQA/HighRS-VQA~\citep{wang2025geollava}
provide the first 8K-scale corpus. Yet question generation in each
is conditioned on image content alone and is agnostic to GSD, so
that a query about ``small vehicles'' may be sampled identically at
$0.5$\,m and $10$\,m even though only the former is physically
answerable. A few efforts acknowledge scale as a supervision axis,
including MMM-RS~\citep{luo2024mmm} for text-to-image generation,
Landsat30-AU~\citep{ma2026landsat30} for coarse-resolution
($\sim$30\,m) captioning, and Scale-MAE~\citep{reed2023scale} at the
representation level, but none conditions the VQA pipeline on a
continuous GSD scalar shared with the model's parameter-activation
mechanism. \textbf{GeoScale-VQA} is the first such corpus, closing
the loop with CS-HLoRA.

\begin{table}[t]
\centering
\caption{Existing assets used in this work and their license or terms-of-use information.}
\label{tab:asset_licenses}
\resizebox{\linewidth}{!}{%
\begin{tabular}{llll}
\toprule
\textbf{Asset} & \textbf{Type} & \textbf{Version / Access Date} & \textbf{License / Terms} \\
\midrule
Qwen3-VL-8B-Instruct
& Model backbone
& HuggingFace model card; accessed 2026-05-06
& Apache-2.0 \\

XLRS-Bench
& Benchmark
& GitHub/HuggingFace release; accessed 2026-05-06
& CC BY-NC-SA 4.0 for repository; HF annotations under CC BY-NC 4.0; images retain source-specific terms \\

OmniEarth-Bench
& Benchmark
& HuggingFace release 2025-05-15; accessed 2026-05-06
& CC BY-NC-SA 4.0 \\

LMMs-Eval
& Evaluation toolkit
& GitHub main branch; accessed 2026-05-06
& MIT for main pipeline code; Apache-2.0 for added multimodal tasks/models \\

RSVQA-HR
& Dataset
& Zenodo v1.0; published 2022-03-10; accessed 2026-05-06
& CC BY 4.0 \\

RSVQA-LR
& Dataset
& Zenodo v1.0; published 2022-03-10; accessed 2026-05-06
& CC BY 4.0 \\

GeoLLaVA-8K / XLRS-related released assets
& Model / dataset / benchmark resource
& GitHub/HuggingFace release; accessed 2026-05-06
& Non-commercial research use; associated XLRS/GeoLLaVA data follow CC BY-NC or source-specific image terms \\

PatternNet
& Dataset
& HuggingFace dataset card; accessed 2026-05-06
& Research purposes only \\

MtSCCD
& Dataset
& National Remote Sensing Bulletin article, 2024; accessed 2026-05-06
& Research/academic use; explicit redistribution license not stated on the article page \\

\bottomrule
\end{tabular}%
}
\end{table}

\section{Limitations}
\label{sec:limitations}

ScaleEarth demonstrates that GSD can serve as a continuous conditioning
variable for remote-sensing VLMs, but several limitations remain. First,
\textsc{GeoScale-VQA} is generated by Qwen3-VL-32B with tier-conditioned
prompts, so its quality is bounded by the generator and by the coverage
of our filtering procedure. Although we apply validation to remove
unverifiable or hallucinated samples, subtle visual ambiguities may still
remain. In addition, the GSD tiers used during data construction, high
resolution below $0.2$\,m, mid resolution between $0.2$ and $1.0$\,m,
and low resolution above $1.0$\,m, follow domain convention rather than
being learned adaptively from data.

Second, our experiments are limited in scale coverage and model family.
The training data spans GSD values from $0.062$\,m to $10$\,m, and we do
not evaluate extreme regimes such as sub-centimeter UAV imagery or coarse
satellite products at $30$\,m resolution and beyond. All experiments are
conducted with Qwen3-VL-8B; although CS-HLoRA only modifies the LoRA
inner space and is therefore compatible with other VLM backbones in
principle, cross-backbone validation remains future work. Finally,
extensions to other domains, such as medical imaging, microscopy, and
autonomous driving, should be interpreted as promising research directions
rather than established empirical claims.

\end{document}